\documentclass{article} % For LaTeX2e
\usepackage{iclr2021_conference,times}

\usepackage{amsmath,amsfonts,bm}

\def\eqref#1{equation~\ref{#1}}
\def\1{\bm{1}}

\def\vx{{\bm{x}}}

\def\vz{{\bm{z}}}

\DeclareMathAlphabet{\mathsfit}{\encodingdefault}{\sfdefault}{m}{sl}
\SetMathAlphabet{\mathsfit}{bold}{\encodingdefault}{\sfdefault}{bx}{n}

\newcommand{\R}{\mathbb{R}}

\usepackage{hyperref}
\usepackage{url}
\usepackage[pdftex]{graphicx}
\usepackage{amsthm}
\usepackage{amsmath}
\usepackage{subfig}

\theoremstyle{definition}
\newtheorem*{definition}{Definition}

\theoremstyle{remark}

\theoremstyle{plain}
\newtheorem{prop}{Proposition}

\title{Very Deep VAEs Generalize Autoregressive Models and Can Outperform Them on Images}

\author{Rewon Child\\
OpenAI\\
San Francisco, CA\\
\texttt{rewon@openai.com} \\
}

\iclrfinalcopy % Uncomment for camera-ready version, but NOT for submission.
\begin{document}

\maketitle

\begin{abstract}

% We present a hierarchical VAE that, for the first time, generates samples quickly \textit{and} gets higher log-likelihood than the PixelCNN on all natural image benchmarks. Traditionally, slow-sampling autoregressive models like the PixelCNN have dominated log-likelihood benchmarks. We show this is counterintuitive, because VAEs are able to implement autoregressive models, as well as numerous more efficient generative models, if made sufficiently deep. Encouraged by this observation, we train VAEs of greater depth than previously explored on CIFAR-10, ImageNet, and FFHQ. We find that, in comparison to the PixelCNN, these very deep VAEs achieve higher likelihoods, use fewer parameters, generate samples thousands of times faster, and are more easily applied to high-resolution images. We visualize the generative process and show the VAEs learn efficient hierarchical visual representations. Our findings suggest previous VAEs were not deep enough. We release our source code and models at \url{https://github.com/openai/vdvae}.

We present a hierarchical VAE that, for the first time, generates samples quickly \textit{and} outperforms the PixelCNN in log-likelihood on all natural image benchmarks. We begin by observing that, in theory, VAEs can actually represent autoregressive models, as well as faster, better models if they exist, when made sufficiently deep. Despite this, autoregressive models have historically outperformed VAEs in log-likelihood. We test if insufficient depth explains why by scaling a VAE to greater stochastic depth than previously explored and evaluating it CIFAR-10, ImageNet, and FFHQ. In comparison to the PixelCNN, these very deep VAEs achieve higher likelihoods, use fewer parameters, generate samples thousands of times faster, and are more easily applied to high-resolution images. Qualitative studies suggest this is because the VAE learns efficient hierarchical visual representations. We release our source code and models at \url{https://github.com/openai/vdvae}.

\end{abstract}

\begin{figure}[h!]
  \centering
  \includegraphics[width=0.9\linewidth]{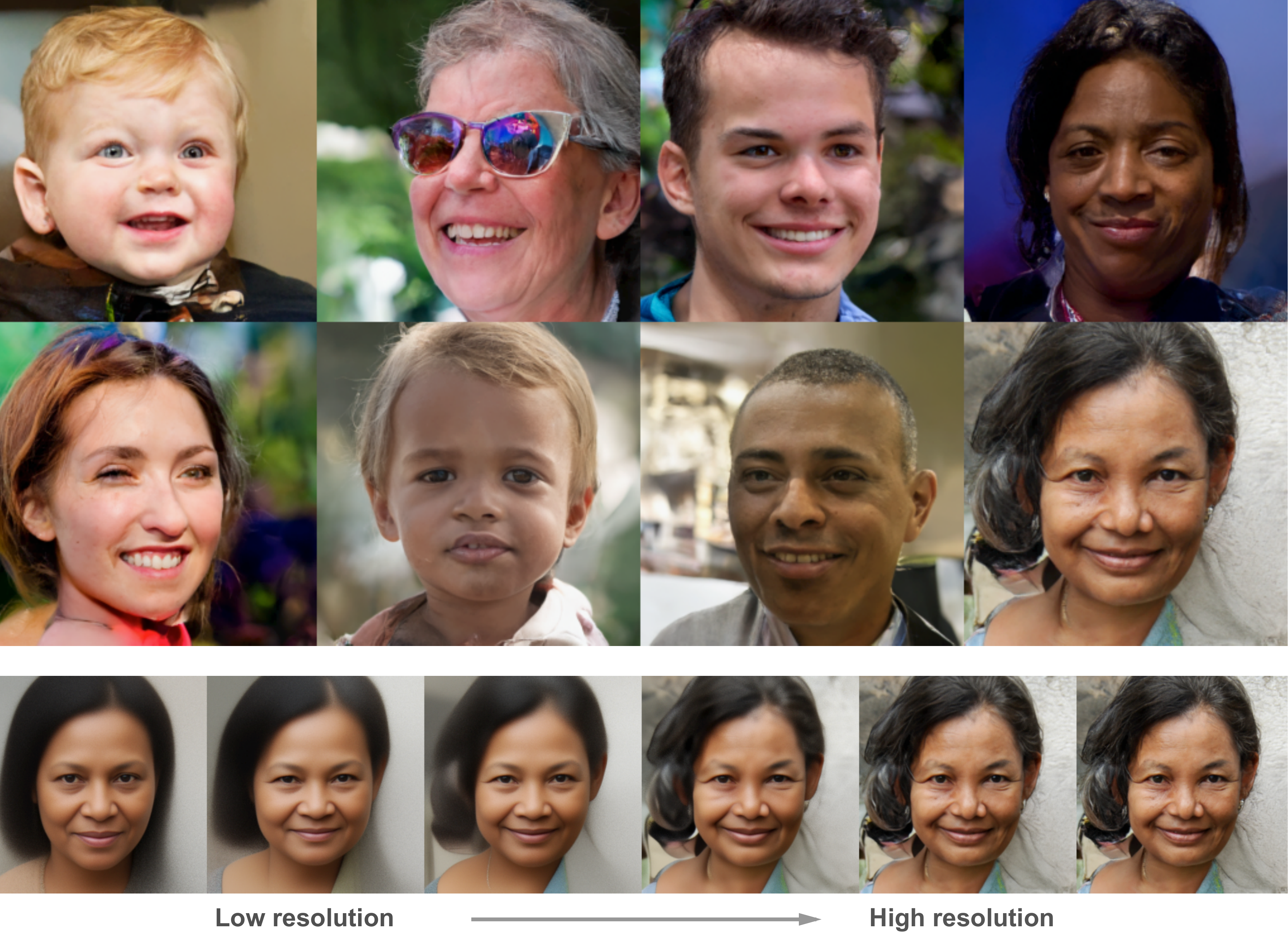}
  \caption{\textbf{Selected samples from our very deep VAE on FFHQ-256, and a demonstration of the learned generative process.} VAEs can learn to first generate global features at low resolution, then fill in local details in parallel at higher resolutions. When made sufficiently deep, this learned, parallel, multiscale generative procedure attains a higher log-likelihood than the PixelCNN.}
\end{figure}

\section{Introduction}

One potential path to increased data-efficiency, generalization, and robustness of machine learning methods is to train generative models. These models can learn useful representations without human supervision by learning to create examples of the data itself. Many types of generative models have flourished in recent years, including likelihood-based generative models, which include autoregressive models \citep{uria2013rnade}, variational autoencoders (VAEs) \citep{kingma2014stochastic,rezende2014stochastic}, and invertible flows \citep{dinh2014nice,dinh2016density}. Their objective, the negative log-likelihood, is equivalent to the KL divergence between the data distribution and the model distribution. A wide variety of models can be compared and assessed along this criteria, which corresponds to how well they fit the data in an information-theoretic sense.

Starting with the PixelCNN \citep{van2016conditional}, autoregressive models have long achieved the highest log-likelihoods across many modalities, despite counterintuitive modeling assumptions. For example, although natural images are observations of latent scenes, autoregressive models learn dependencies solely between observed variables. That process can require complex function approximators that integrate long-range dependencies \citep{oord2016wavenet,child2019generating}. In contrast, VAEs and invertible flows incorporate latent variables and can thus, in principle, learn a simpler model that mirrors how images are actually generated. Despite this theoretical advantage, on the landmark ImageNet density estimation benchmark, the Gated PixelCNN still achieves higher likelihoods than all flows and VAEs, corresponding to a better fit with the data.

Is the autoregressive modeling assumption actually a better inductive bias for images, or can VAEs, sufficiently improved, outperform autoregressive models? The answer has significant practical stakes, because large, compute-intensive autoregressive models \citep{strubell2019energy} are increasingly used for a variety of applications \citep{oord2016wavenet,brown2020language,dhariwal2020jukebox,chen2020generative}. Unlike autoregressive models, latent variable models only need to learn dependencies between latent and observed variables; such models can not only support faster synthesis and higher-dimensional data, but may also do so using smaller, less powerful architectures.

% Latent variable models offer the promise of fast synthesis and greater scalability to high-dimensional data, and fundamentally only need to model dependencies between latent variables and observed variables, which may be achievable with smaller and less powerful models.

We start this work with a simple but (to the best of our knowledge) unstated observation: hierarchical VAEs \textit{should} be able to at least match autoregressive models, because autoregressive models are equivalent to VAEs with a powerful prior and restricted approximate posterior (which merely outputs observed variables). In the worst case, VAEs should be able to replicate the functionality of autoregressive models; in the best case, they should be able to learn better latent representations, possibly with much fewer layers, if such representations exist.

We formalize this observation in Section \ref{depthvae}, showing it is only true for VAEs with more stochastic layers than previous work has explored. Then we experimentally test it on competitive natural image benchmarks. Our contributions are the following:

% In section 2, we formalize this observation and show that it is only true for VAEs with many more stochastic layers than previous work has explored. In section 3, we introduce modifications we made to hierarchical VAEs to successfully train them at greater depths, and in section 4 we quantitatively assess the performance of deeper VAEs, finding they can outperform autoregressive networks. We then qualitatively study the models, finding that they are more parameter efficient, easily scaled to higher resolution data (like 1024x1024 FFHQ), and may require fewer long-term dependencies than autoregressive models. Our contributions are the following:

\begin{itemize}
\item We provide theoretical justification for why greater depth (up to the data dimension $D$, but also as low as some value $K \ll D$) could improve VAE performance (Section \ref{depthvae})
\item We introduce an architecture capable of scaling past 70 layers, when previous work explored at most 30 (Section \ref{archtext})
\item We verify that depth, independent of model capacity, improves log-likelihood, and allows VAEs to outperform the PixelCNN on all benchmarks (Section \ref{depthexp})
\item Compared to the PixelCNN, we show the model also uses fewer parameters, generates samples thousands of times more quickly, and can be scaled to larger images. We show evidence these qualities may emerge from the model learning an efficient hierarchical representation of images (Section \ref{hierarchyexp}) 
\item We release code and models at \url{https://github.com/openai/vdvae}.
\end{itemize}

\begin{figure}[t]
  \centering
  \includegraphics[width=1.0\linewidth]{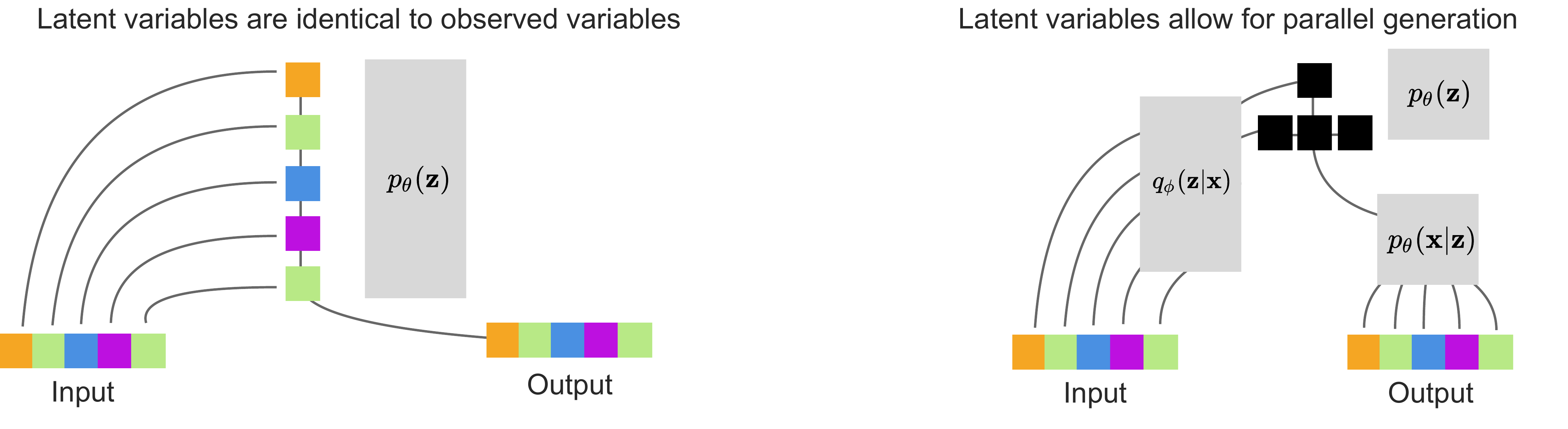}
  \caption{\textbf{Different possible learned generative models in a VAE.} \textbf{Left:} A hierarchical VAE can learn an autoregressive model by using the deterministic identity function as an encoder, and learning the autoregression in the prior. \textbf{Right:} Learning the encoder can lead to efficient hierarchies of latent variables (black). If the bottom group of three latent variables is conditionally independent given the first, they can be generated in parallel within a single layer, potentially leading to faster sampling.} 
  \label{fig:arequiv}
\end{figure}

\section{Preliminaries}
We review prior work and introduce some of the basic terminology used in the field.

\subsection{Variational Autoencoders}
Variational autoencoders \citep{kingma2014stochastic,rezende2014stochastic} consist of a \textit{generator} $p_\theta(\vx | \vz)$, a \textit{prior} $p_\theta(\vz)$, and an \textit{approximate posterior} $q_\phi(\mathbf{z} | \mathbf{x})$. Neural networks $\phi$ and $\theta$ are trained end-to-end with backpropagation and the reparameterization trick in order to maximize the evidence lower bound (ELBO):

\begin{equation}
\log p_\theta(\mathbf{x}) \ge E_{\mathbf{z} \sim q_\phi(\mathbf{z} | \mathbf{x})} \log p_\theta(\mathbf{x}|\mathbf{z}) - D_{KL}[q_\phi(\mathbf{z} | \mathbf{x}) || p_\theta(\mathbf{z})]
\label{elbo}
\end{equation}

See \citet{kingma2019introduction} for an in-depth introduction. There are many choices for what networks are used for $p_\theta (\vx | \vz)$, $q_\phi( \vz | \vx)$, and whether $p_\theta(\vz)$ is also learned or set to a simple distribution. 

We study VAEs with independent $p_\theta (\vx | \vz)$ -- that is, where each observed $x_i$ is output without conditioning on any other $x_j$. This ensures generation time does not increase linearly with the dimensionality of the data, and requires that these VAEs learn to incorporate the complexity of the data into a rich distribution over latent variables $\vz$. It is possible to have autoregressive $p_\theta (\vx | \vz)$ \citep{gulrajani2016pixelvae}, but generation is slow for these models. They also sometimes ignore latent variables entirely, becoming equivalent to normal autoregressive models (\cite{chen2016variational}). 

\subsection{Hierarchical Variational Autoencoders}
\label{hvae}
Much of the early work on VAEs incorporate fully-factorized Gaussian $q_\phi( \vz | \vx)$ and $p_\theta(\vz)$. This can lead to poor outcomes if the latent variables required for good generation take on a more complex distribution, as is common with independent $p_\theta(\vx|\vz)$. One of the simplest methods of gaining greater expressivity in both distributions is to use a hierarchical VAE, which has several \textit{stochastic layers} of latent variables. These variables are emitted in groups $\vz_0, \vz_1, ..., \vz_N$, which are conditionally dependent upon each other in some way. For images, latent variables are typically output in feature maps of varying resolutions, with $\vz_0$ corresponding to a small number of latent variables at low resolution at the ``top" of the network, and $\vz_N$ corresponding to a larger number of latent variables at high resolution at the ``bottom".

One particularly elegant conditioning structure is the \textit{top-down VAE}, introduced in \cite{sonderby2016ladder}. In this model, both the prior and the approximate posterior generate latent variables in the same order:

\begin{equation}
p_\theta( \vz) = p_\theta( \vz_0) p_\theta( \vz_1 | \vz_0) ... p_\theta( \vz_N | \vz_{<N})
\label{topdownpr}
\end{equation}
\begin{equation}
q_\phi( \vz | \vx) = q_\phi( \vz_0 | \vx) q_\phi( \vz_1 | \vz_0, \vx) ... q_\phi( \vz_N | \vz_{<N}, \vx)
\label{topdownpo}
\end{equation}

A diagram of this process appears in Figure 3. A typical implementation of this model has $\phi$ first perform a deterministic ``bottom-up" pass on the data to generate features, then processes the groups of latent variables from top to bottom, using feedforward networks to generate features which are shared between the approximate posterior, prior, and reconstruction network $p_\theta(\vx|\vz)$. We adopt this base architecture as it is simple, empirically effective, and has been postulated to resemble biological processes of perception \citep{dayan1995helmholtz}.

\begin{figure}[t]
  \centering
  \includegraphics[width=1.0\linewidth]{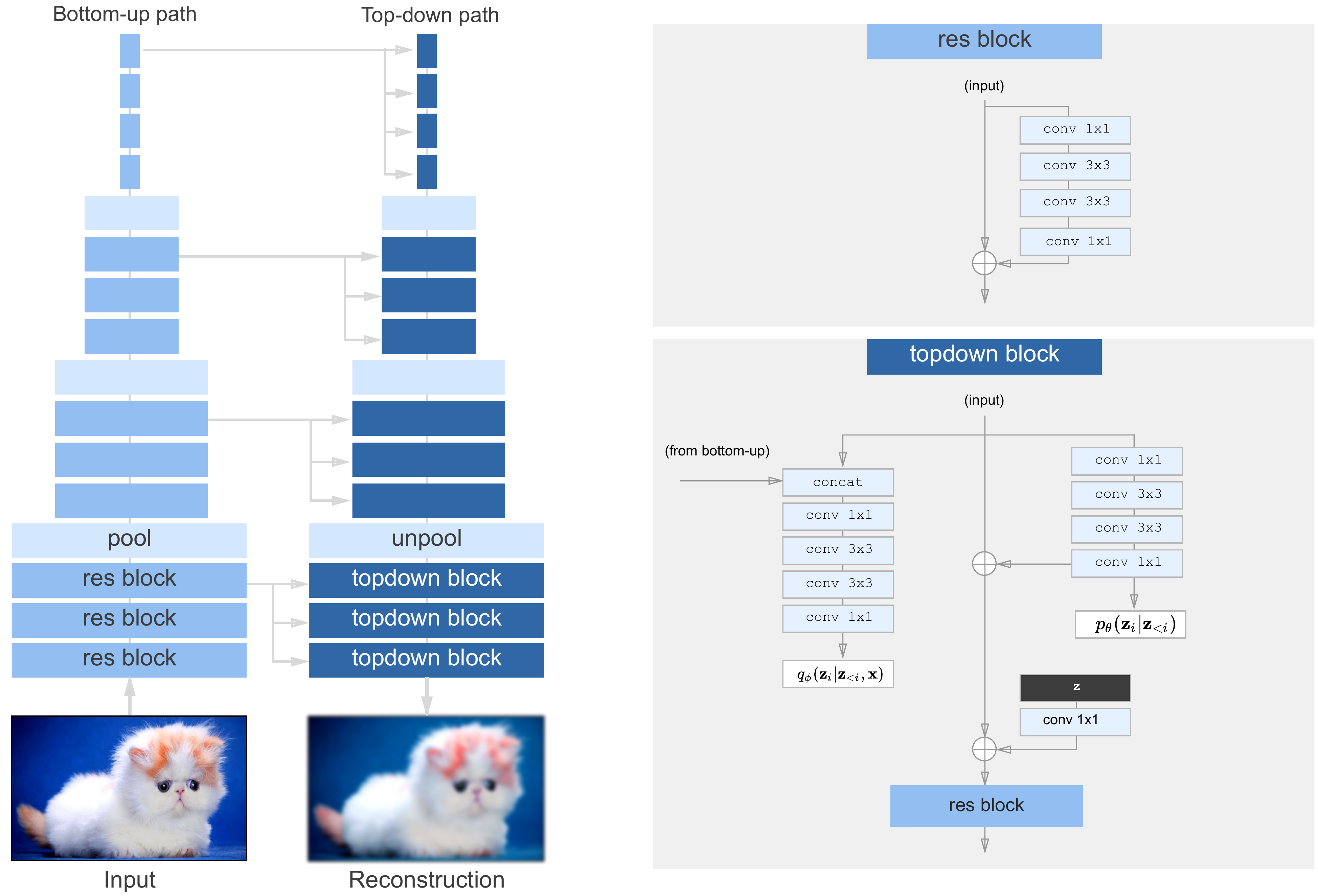}
  \caption{\textbf{A diagram of our top-down VAE architecture.} Residual blocks are similar to bottleneck ResNet blocks \citep{he2016deep}. Each convolution is preceded by the GELU nonlinearity \citep{hendrycks2016gaussian}. $q_\phi(.)$ and $p_\theta(.)$ are diagonal Gaussian distributions. $\mathbf{z}$ is sampled from $q_\phi(.)$ during training, and $p_\theta(.)$ when sampling. We use average pooling and nearest-neighbor upsampling for pool and unpool layers.} 
  \label{fig:arch}
\end{figure}

\section{Why depth matters for hierarchical VAEs}
\label{depthvae}
We find that hierarchical VAEs with sufficient depth can not only learn arbitrary orderings over observed variables, but also learn more effective latent variable distributions, if such distributions exist. We present these results below.

 % hierarchical VAEs can outperform autoregressive networks when they have many stochastic layers. We do this by showing VAEs much deeper than previous work can not only implement autoregressive models, but also a much larger class of generators which are potentially more efficient.

\theoremstyle{definition}
\begin{definition}[$N$-layer VAE]
A deep hierarchical VAE with $N$ stochastic layers, independent $p(\vx|\vz)$, and the top-down factorization of the prior and approximate posterior in Equations \ref{topdownpr}-\ref{topdownpo}.
\end{definition}

% All proofs we include in the Appendix.

\theoremstyle{plain}
\begin{prop}N-layer VAEs generalize autoregressive models when N is the data dimension
\end{prop}

\begin{prop}$N$-layer VAEs are universal approximators of $N$-dimensional latent densities
\end{prop}

Proposition 1 (proof in Appendix, also visualized in Figure \ref{fig:arequiv}, left) leads to a possible explanation of why autoregressive models to date have outperformed VAEs: \textit{they are deeper}, in the sense of statistical dependence. A VAE must be as deep as the data dimension $D$ (3072 layers in the case of 32x32 images) if the images truly require $D$ steps to generate. 

Luckily, however, Proposition 2 (proof and further technical requirements in Appendix) suggests that shorter procedures, if they exist, are also learnable. $N=D$ is an extreme case, where the most effective latent variables $\vz \in \R^D$ may simply be copies of the observed variables. But if for some $K < D$ there exist latent variables $\vz \in \R^K$ that the generator can use to more efficiently compress the data, Proposition 2 states a $K$-layer VAE can learn the posterior and prior distribution over those variables.

Such shorter generative paths could emerge in two ways. First, as depicted in Figure \ref{fig:arequiv} (right), if the model discovers that certain variables are conditionally independent given others, the model can generate them in parallel inside a single layer, where $q_\phi(\vz_N | \vz_{<N}, \vx) = \prod_d q_\phi(z_{N}^{(d)}| \vz_{<N}, \vx)$. We hypothesize these efficient hierarchies should emerge in images, as they contain many spatially independent textures, and study this in Section \ref{hierarchyexp}. Second, the model could learn a low-dimensional representation of the data. \cite{dai2019diagnosing} recently showed that when a VAE is trained on data distributed on a $K$-dimensional manifold embedded in $\R^D$, a VAE will only activate $K$ dimensions in its latent space, meaning that the VAE will require fewer layers unless the manifold dimension is $D$, which is unlikely to be the case for images.

It is difficult to ascertain the lowest possible value of $K$ for a given dataset, but it may be deeper than most hierarchical VAEs to date. Images have many thousands of observed variables, but early hierarchical VAEs did not exceed 3 layers, until \cite{maaloe2019biva} investigated a Gaussian VAE with 15 layers and found it displayed impressive performance along a variety of measures. \cite{kingma2016improved} and \cite{vahdat2020nvae} additionally explored networks up to 12 and 30 layers. (These additionally incorporated additional statistical dependencies in the approximate posterior through the usage of inverse autoregressive flow \citep{kingma2016improved}, an alternative approach which we contrast with our approach in Section \ref{iaf}). Nevertheless, given these results we hypothesize that greater depth may improve the performance of VAEs. In the next section, we introduce an architecture capable of scaling to a greater number of stochastic layers. In Section \ref{depthexp} we show depth indeed improves performance.

\section{An architecture for very deep VAEs}
\label{archtext}

We consider a ``very deep" VAE to simply be one with greater depth than has previously been explored (and do not define it to be a specific number of layers). As existing implementations of VAEs did not support many more stochastic layers than they were trained on, we reimplemented a minimal VAE with the sole aim of increasing the number of stochastic layers. This VAE consists only of convolutions, nonlinearities, and Gaussian stochastic layers. It does not exhibit posterior collapse even for large numbers of stochastic layers. We describe key architectural choices here and refer readers to our source code for more details.

\subsection{Architectural components and initialization}
A diagram of our network appears in Figure \ref{fig:arch}. It resembles the ResNet VAE in \cite{kingma2016improved}, but with bottleneck residual blocks. For each stochastic layer, the prior and posterior are diagonal Gaussian distributions, as used in prior work \citep{maaloe2019biva}.

As an alternative to weight normalization and data-dependent initialization \citep{salimans2016weight}, we adopt the default PyTorch weight intialization. The one exception is the final convolutional layer in each residual bottleneck block, which we scale by $\frac{1}{\sqrt{N}}$, where N is the depth (similar to \cite{radford2019language,child2019generating,zhang2019fixup}). This residual scaling improves stability and performance with many layers, as we show in the Appendix (Table \ref{abl:residualscaling}).

Additionally, we use nearest-neighbor upsampling for our ``unpool" layer, which when paired with our ResNet architecture, allows us to completely remove the ``free bits" and KL ``warming up" terms that appear in related work. As we detail in the Appendix (Figure \ref{abl:postcollapse}), when upsampling is done through transposed convolutional layer, the network may ignore layers at low resolution (for instance, 1x1 or 4x4 layers). We found no evidence of posterior collapse in any networks trained with nearest neighbor interpolation.

\subsection{Stabilizing training with gradient skipping}
VAEs have notorious ``optimization difficulties," which are not frequently discussed in the literature but nevertheless well-known by practitioners. These manifest as extremely high reconstruction or KL losses and corresponding large gradient norms (up to $1e15$). We address this by skipping updates with a gradient norm above a certain threshold, set by hyperparameter. Though we select high thresholds that affect fewer than 0.01\% of updates, this technique almost entirely eliminates divergence, and allows networks to train smoothly. We plot the evolution of grad norms and the values we select in (Figure \ref{abl:gradskips}). An alternative approach to stabilizing networks may be the spectral regularization method introduced in \cite{vahdat2020nvae}.

\begin{table}
  \caption{\textbf{Loss by network with different configurations of stochastic layers on ImageNet-32} (similar trends appear on CIFAR-10). \textbf{Left}: Networks with equal number of layers, but with lower stochastic depth as described in Section \ref{depthexp}. Increasing depth up to 48 layers still shows gains, which is farther than previous work has explored. \textbf{Right}: Networks with 48 layers, but distributed at different resolutions. We find higher resolutions benefit more from layers.}
  \label{tbl:ablations}
  \centering
  \begin{tabular}{ccc}
    \textbf{Depth} & \textbf{Params} & \textbf{Test Loss}\\
    \hline
    3 & 41M &  4.30 \\
    6 & 41M & 4.18 \\
    12 & 41M & 4.06 \\
    24 & 41M & 3.98 \\
    48 & 41M & 3.95 \\
  \end{tabular}
  \quad
  \quad
  \quad
  \quad
  \begin{tabular}{cc}
    \textbf{Distribution of 48 layers} & \textbf{Test Loss}\\
    32x32 \, 16x16 \, 8x8 \, 4x4 \, 1x1 \, \\
    \hline
    10 \, \, \, \, \, 10 \, \, \, 10 \, \,  10 \,  \, 8  & 3.98 \\
    12 \, \, \, \, \, 12 \, \, \, 10 \, \,  8 \, \,  \, 6  & 3.97 \\
    14 \, \, \, \, \, 14 \, \, \, 10 \, \,  6 \, \,  \, 4  & 3.96 \\
    16 \, \, \, \, \, 16 \, \, \, 10 \, \,  4 \, \,  \, 2  & 3.95 \\
  \end{tabular}
\end{table}

\section{Experiments}
We trained very deep VAEs on challenging natural image datasets. All hyperparameters for experiments are available in the Appendix and in our source code.

\subsection{Statistical depth, independent of capacity, improves performance}
\label{depthexp}
We first tested whether greater statistical depth, independent of other factors, can result in improved performance. We trained a network with 48 layers for 600k steps on ImageNet-32, grouping layers to output variables independently instead of conditioning on each other. If the input for the $i$th topdown block is $x_i$, we can make $K$ consecutive blocks independent by setting $x_{i+1}$, ..., $x_{i+K}$ all equal to $x_i$. (Normally, $x_{i+1} = x_i + f(\mathrm{block}(x_i))$). This technique reduces the stochastic depth without affecting parameter count. Stochastic depth shows a clear correlation with performance, even up to 48 layers, which is past what previous work has explored (Table \ref{tbl:ablations}, left).

We then tested our hypothesis at scale. We trained networks on CIFAR-10, ImageNet-32, and ImageNet-64 with greater numbers of stochastic layers, but with \textit{fewer} parameters than related work (see Table \ref{tbl:likelihood}). On CIFAR-10, we trained a model with 45 stochastic layers and only 39M parameters, and found it achieved a test log-likelihood of 2.87 bits per dim (average of 4 seeds). On ImageNet-32 and ImageNet-64, we trained networks with 78 and 75 stochastic layers and only approximately 120M parameters, and achieved likelihoods of 3.80 and 3.52.

On all tasks, these results outperform all GatedPixelCNN/PixelCNN++ models, and all non-autoregressive models, while using similar or fewer parameters. These results support our hypothesis that \textit{stochastic depth}, as opposed to other factors, explains the gap between VAEs and autoregressive models.

\subsection{Very deep VAEs learn an efficient hierarchical ordering}
\label{hierarchyexp}
\begin{figure}[t]
  % \centering
  \includegraphics[width=1.0\linewidth]{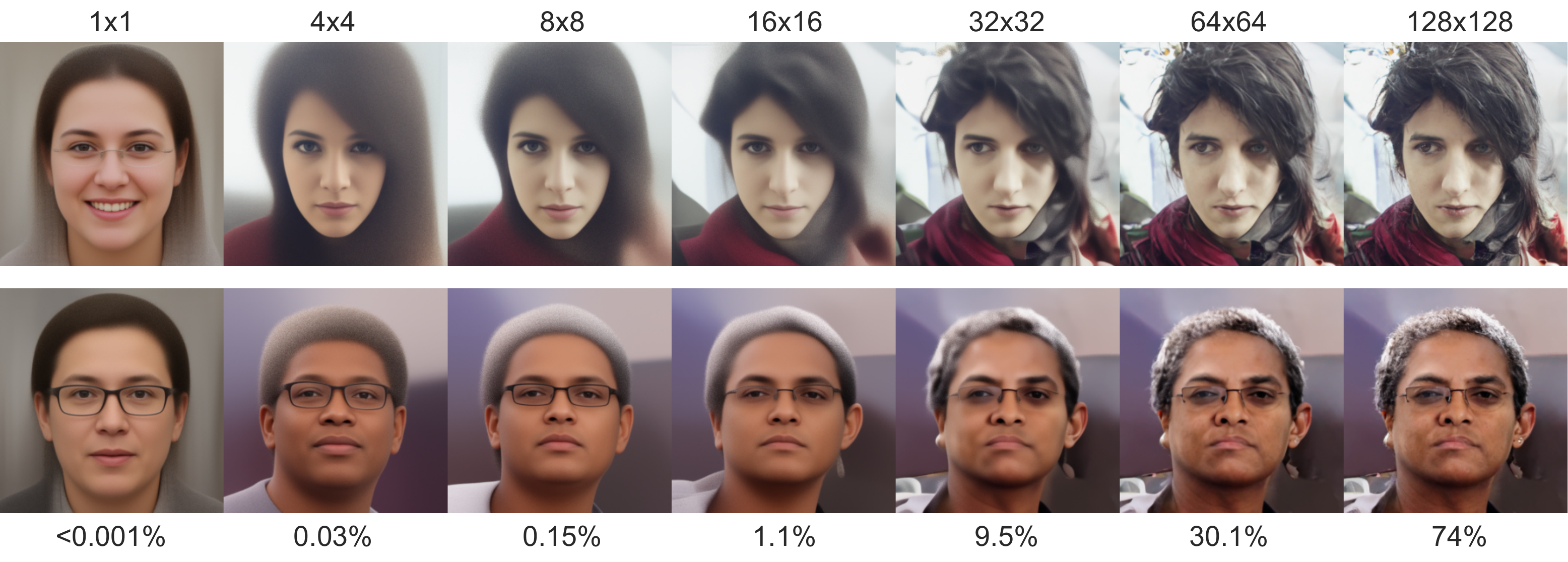}
  \caption{\textbf{Cumulative percentage of latent variables at a given resolution, and reconstructions of samples on FFHQ-256.} We sample latent variables from the approximate posterior until the given resolution, and sample the rest from the prior at low temperature. This shows what images are likely given a subset of latent variables. Low-resolution latents comprise a small fraction of the total latents, but encode significant portions of the global structure. This suggests deep VAEs learn efficient, hierarchical representations of the data.} 
  \label{fig:latents}
\end{figure}

One question that emerges from the analysis in Section 3 is whether VAEs need to be as deep as autoregressive models, or whether they can learn a latent hierarchy of conditionally independent variables which are able to be synthesized in parallel. We qualitatively show this is true in Figure \ref{fig:latents}. For FFHQ-256 images, the first several layers at low resolution almost wholly determine the global features of the image, even though they only account for less than 1\% of the latent variables. The rest of the high-resolution variables appear to be spatially independent, meaning they can be emitted in parallel in a number of layers much lower than the dimensionality of the image. This efficient hierarchical representation may underlie the VAE's ability to achieve better log-likelihoods than the PixelCNN while simultaneously sampling thousands of times faster. This can be viewed as a \textit{learned} parallel multiscale generation method, unlike the handcrafted approaches of \cite{kolesnikov2017pixelcnn,menick2018generating,reed2017parallel}.

Additionally, we found that on all datasets we tested, very deep VAEs used roughly 30\% fewer parameters than the PixelCNN (Table \ref{tbl:likelihood}). One possible explanation is that the learned hierarchical generation procedure involves fewer long-range dependencies, or may otherwise be simpler to learn. 

We found that networks in general benefited from more layers at higher resolutions (Table \ref{tbl:ablations}, right). This suggests that global features may account for a smaller fraction of information than local details and textures, and that it is important to have many latent variables at high resolution.

\subsubsection{Very deep VAEs are easily scaled to high dimensional data}
\label{scaleexp}
Scaling autoregressive models to higher resolutions presents several challenges. First, the sampling time and memory requirements of autoregressive models increase \textit{linearly} with resolution. This scaling makes datasets like FFHQ-256 and FFHQ-1024 intractable for naive approaches. Although clever factorization techniques have been adopted for 256x256 images \citep{menick2018generating}, such factorizations may not be as effective for alternate datasets or higher-resolution images.

Our VAE, in contrast, readily scales to higher resolutions. The same network used for 32x32 images can be applied to 1024x1024 images by introducing a greater number of upsampling layers throughout the network. We found we could train an equal number of steps (1.5M) using a similar number of training resources (32 GPUs for 2.5 weeks) on both 32x32 and 1024x1024 images with few hyperparameter changes (see Appendix for hyperparameters). Samples from both models (displayed in Appendix) require a single forward pass of the model to generate, with only minor differences in runtime. An autoregressive model, on the other hand, would require a thousand times more network evaluations to sample 1024x1024 images and likely require a custom training procedure.

\begin{table}[t!]
  \caption{\textbf{Our main results on standard benchmark datasets.} Very deep VAEs outperform PixelCNN-based autoregressive models with fewer parameters while maintaining fast sampling. ``Depth" refers to the number of stochastic layers for hierarchical VAEs (although BIVA and IAF-based networks have additional statistical dependencies). Sampling refers to the number of network evaluations per sample, and $D$ designates the dimensionality of the data. An asterisk ($^*$) denotes our estimate of parameters. Samples for ImageNet and CIFAR-10 are in the Appendix.}
  \label{tbl:likelihood}
  \centering
  \begin{tabular}{llllll}
    & Model type & Params & Depth & Sampling & NLL \\
    \hline
    \\
    \textbf{CIFAR-10}\\
    PixelCNN++ \citep{salimans2017pixelcnn++} & AR & 53M$^*$ &  & $D$ & 2.92 \\
    PixelSNAIL \citep{chen2017pixelsnail} & AR & &  & $D$ & 2.85\\
    Sparse Transformer \citep{child2019generating} & AR & 59M &  & $D$ & \textbf{2.80}\\
    VLAE \citep{chen2016variational}  & VAE & &  & $D$ & $\le$ 2.95\\
    IAF-VAE \citep{kingma2016improved}  & VAE & & 12 & $1$ & $\le$ 3.11\\
    % Glow  & Flow & No & 3.11\\
    Flow++ \citep{ho2019flow++} & Flow & 31M &  & $1$ & $\le$ 3.08\\
    BIVA \citep{maaloe2019biva}  & VAE & 103M & 15 & $1$ & $\le$ 3.08\\
    NVAE \citep{vahdat2020nvae}  & VAE & 131M & 30 & $1$ & $\le$ 2.91\\
    Very Deep VAE \textbf{(ours)} & VAE & 39M & 45 & $1$ & $\le$ \textbf{2.87}\\
    \\
    \textbf{ImageNet-32}\\
    Gated PixelCNN & AR & 177M$^*$ & 10  & $D$ & 3.83\\
    Image Transformer \citep{parmar2018image} & AR &  & & $D$ & \textbf{3.77}\\
    BIVA  & VAE & 103M$^*$ & 15  & $1$ & $\le$ 3.96\\
    NVAE  & VAE & 268M & 28  & $1$ & $\le$ 3.92\\
    Flow++ & Flow & 169M  & & $1$ & $\le$ 3.86\\
    Very Deep VAE \textbf{(ours)} & VAE & 119M & 78 & $1$ & $\le$ \textbf{3.80}\\
    \\
    \textbf{ImageNet-64}\\
    Gated PixelCNN & AR & 177M$^*$ &  & $D$ & 3.57\\
    SPN \citep{menick2018generating} & AR & 150M  & & $D$ & 3.52\\
    Sparse Transformer & AR & 152M  & & $D$ & \textbf{3.44}\\
    Glow \citep{kingma2018glow}  & Flow &  & & $1$ & 3.81\\
    Flow++ & Flow & 73M  & & $1$ & $\le$ 3.69\\
    Very Deep VAE \textbf{(ours)} & VAE & 125M & 75 & $1$ & $\le$ \textbf{3.52}\\
    \\
    \textbf{FFHQ-256 (5 bit)}\\
    NVAE  & VAE & & 36 & $1$ & $\le$ 0.68\\
    Very Deep VAE \textbf{(ours)} & VAE & 115M & 62 & $1$ & $\le$ \textbf{0.61}\\
    \\
    \textbf{FFHQ-1024 (8 bit)}\\
    Very Deep VAE \textbf{(ours)} & VAE & 115M & 72 & $1$ & $\le$ \textbf{2.42}\\
  \end{tabular}
\end{table}

% \subsubsection{Learning a dynamic ordering}
% We also found that layers were more useful at higher resolutions than at lower resolutions (Table \ref{tbl:ablations}, right). As far as we know, this architectural detail has not been noted in prior work, and suggests that increased conditioning for fine-grained details is helpful. This is one practical approach to overcoming the advantage of autoregressive models in modeling textures and local details, which can account for the majority of bits in an image.

\section{Related work and discussion}

Our work is inspired by previous and concurrent work in hierarchical VAEs \citep{sonderby2016ladder,maaloe2019biva,vahdat2020nvae}. Relative to these works, we provide some justification for why deeper networks may perform better, introduce a new architecture, and empirically demonstrate gains in log-likelihood. Many aspects of prior work are complementary with ours and could be combined. \cite{maaloe2019biva}, for instance, incorporates a ``bottom-up" stochastic path that doubles the depth of the approximate posterior, and \cite{vahdat2020nvae} introduces a number of powerful architecture components and improved training techniques. We seek here not to introduce a significantly better method than these alternatives, but to demonstrate that depth is a key overlooked factor in most prior approaches to VAEs.

Diffusion models can be seen as deep VAEs that, like autoregressive models, have a specific analytical posterior. \cite{ho2020denoising} showed that such models achieve impressive sample quality with great depth, which is in line with our observations that greater depth is helpful for VAEs. One benefit of the VAEs we outline in this work over diffusion models is that our VAEs generate samples with a single network evaluation, whereas diffusion models currently require a large number of network evaluations per sample.

Inverse autoregressive flows (IAF) are also closely related, and we discuss the differences with hierarchical models in Section \ref{iaf}. The work of \cite{zhao2017learning} may also appear to contradict our findings, and we discuss that work in Section \ref{zhao}.

\section{Conclusion}
We argue deeper VAEs should perform better, introduce a deeper architecture, and show it outperforms all PixelCNN-based autoregressive models in likelihood while being more efficient. We hope this encourages work in further improving VAEs and latent variable models.

\subsubsection*{Acknowledgments}
We thank Aditya Ramesh, Pranav Shyam, Johannes Otterbach, Heewoo Jun, Mark Chen, Prafulla Dhariwal, Alec Radford, Yura Burda, Bowen Baker, Raul Puri, and Ilya Sutskever for helpful discussions. We also thank the anonymous reviewers for helping improve our work.

\bibliography{refs}

\begin{thebibliography}{39}
\providecommand{\natexlab}[1]{#1}
\providecommand{\url}[1]{\texttt{#1}}
\expandafter\ifx\csname urlstyle\endcsname\relax
  \providecommand{\doi}[1]{doi: #1}\else
  \providecommand{\doi}{doi: \begingroup \urlstyle{rm}\Url}\fi

\bibitem[Brown et~al.(2020)Brown, Mann, Ryder, Subbiah, Kaplan, Dhariwal,
  Neelakantan, Shyam, Sastry, Askell, et~al.]{brown2020language}
Tom~B Brown, Benjamin Mann, Nick Ryder, Melanie Subbiah, Jared Kaplan, Prafulla
  Dhariwal, Arvind Neelakantan, Pranav Shyam, Girish Sastry, Amanda Askell,
  et~al.
\newblock Language models are few-shot learners.
\newblock \emph{arXiv preprint arXiv:2005.14165}, 2020.

\bibitem[Chen et~al.(2020)Chen, Radford, Child, Wu, Jun, Dhariwal, Luan, and
  Sutskever]{chen2020generative}
Mark Chen, Alec Radford, Rewon Child, Jeff Wu, Heewoo Jun, Prafulla Dhariwal,
  David Luan, and Ilya Sutskever.
\newblock Generative pretraining from pixels.
\newblock In \emph{Proceedings of the 37th International Conference on Machine
  Learning}, 2020.

\bibitem[Chen et~al.(2016)Chen, Kingma, Salimans, Duan, Dhariwal, Schulman,
  Sutskever, and Abbeel]{chen2016variational}
Xi~Chen, Diederik~P Kingma, Tim Salimans, Yan Duan, Prafulla Dhariwal, John
  Schulman, Ilya Sutskever, and Pieter Abbeel.
\newblock Variational lossy autoencoder.
\newblock \emph{arXiv preprint arXiv:1611.02731}, 2016.

\bibitem[Chen et~al.(2017)Chen, Mishra, Rohaninejad, and
  Abbeel]{chen2017pixelsnail}
Xi~Chen, Nikhil Mishra, Mostafa Rohaninejad, and Pieter Abbeel.
\newblock Pixelsnail: An improved autoregressive generative model.
\newblock \emph{arXiv preprint arXiv:1712.09763}, 2017.

\bibitem[Child et~al.(2019)Child, Gray, Radford, and
  Sutskever]{child2019generating}
Rewon Child, Scott Gray, Alec Radford, and Ilya Sutskever.
\newblock Generating long sequences with sparse transformers.
\newblock \emph{arXiv preprint arXiv:1904.10509}, 2019.

\bibitem[Dai \& Wipf(2019)Dai and Wipf]{dai2019diagnosing}
Bin Dai and David Wipf.
\newblock Diagnosing and enhancing vae models.
\newblock \emph{arXiv preprint arXiv:1903.05789}, 2019.

\bibitem[Dayan et~al.(1995)Dayan, Hinton, Neal, and Zemel]{dayan1995helmholtz}
Peter Dayan, Geoffrey~E Hinton, Radford~M Neal, and Richard~S Zemel.
\newblock The helmholtz machine.
\newblock \emph{Neural computation}, 7\penalty0 (5):\penalty0 889--904, 1995.

\bibitem[Dhariwal et~al.(2020)Dhariwal, Jun, Payne, Kim, Radford, and
  Sutskever]{dhariwal2020jukebox}
Prafulla Dhariwal, Heewoo Jun, Christine Payne, Jong~Wook Kim, Alec Radford,
  and Ilya Sutskever.
\newblock Jukebox: A generative model for music.
\newblock \emph{arXiv preprint arXiv:2005.00341}, 2020.

\bibitem[Dinh et~al.(2014)Dinh, Krueger, and Bengio]{dinh2014nice}
Laurent Dinh, David Krueger, and Yoshua Bengio.
\newblock Nice: Non-linear independent components estimation.
\newblock \emph{arXiv preprint arXiv:1410.8516}, 2014.

\bibitem[Dinh et~al.(2016)Dinh, Sohl-Dickstein, and Bengio]{dinh2016density}
Laurent Dinh, Jascha Sohl-Dickstein, and Samy Bengio.
\newblock Density estimation using real nvp.
\newblock \emph{arXiv preprint arXiv:1605.08803}, 2016.

\bibitem[Gulrajani et~al.(2016)Gulrajani, Kumar, Ahmed, Taiga, Visin, Vazquez,
  and Courville]{gulrajani2016pixelvae}
Ishaan Gulrajani, Kundan Kumar, Faruk Ahmed, Adrien~Ali Taiga, Francesco Visin,
  David Vazquez, and Aaron Courville.
\newblock Pixelvae: A latent variable model for natural images.
\newblock \emph{arXiv preprint arXiv:1611.05013}, 2016.

\bibitem[He et~al.(2016)He, Zhang, Ren, and Sun]{he2016deep}
Kaiming He, Xiangyu Zhang, Shaoqing Ren, and Jian Sun.
\newblock Deep residual learning for image recognition.
\newblock In \emph{Proceedings of the IEEE conference on computer vision and
  pattern recognition}, pp.\  770--778, 2016.

\bibitem[Hendrycks \& Gimpel(2016)Hendrycks and Gimpel]{hendrycks2016gaussian}
Dan Hendrycks and Kevin Gimpel.
\newblock Gaussian error linear units (gelus).
\newblock \emph{arXiv preprint arXiv:1606.08415}, 2016.

\bibitem[Ho et~al.(2019)Ho, Chen, Srinivas, Duan, and Abbeel]{ho2019flow++}
Jonathan Ho, Xi~Chen, Aravind Srinivas, Yan Duan, and Pieter Abbeel.
\newblock Flow++: Improving flow-based generative models with variational
  dequantization and architecture design.
\newblock \emph{arXiv preprint arXiv:1902.00275}, 2019.

\bibitem[Ho et~al.(2020)Ho, Jain, and Abbeel]{ho2020denoising}
Jonathan Ho, Ajay Jain, and Pieter Abbeel.
\newblock Denoising diffusion probabilistic models.
\newblock \emph{arXiv preprint arxiv:2006.11239}, 2020.

\bibitem[Huang et~al.(2017)Huang, Touati, Dinh, Drozdzal, Havaei, Charlin, and
  Courville]{huang2017learnable}
Chin-Wei Huang, Ahmed Touati, Laurent Dinh, Michal Drozdzal, Mohammad Havaei,
  Laurent Charlin, and Aaron Courville.
\newblock Learnable explicit density for continuous latent space and
  variational inference.
\newblock \emph{arXiv preprint arXiv:1710.02248}, 2017.

\bibitem[Huang et~al.(2018)Huang, Krueger, Lacoste, and
  Courville]{huang2018neural}
Chin-Wei Huang, David Krueger, Alexandre Lacoste, and Aaron Courville.
\newblock Neural autoregressive flows.
\newblock \emph{arXiv preprint arXiv:1804.00779}, 2018.

\bibitem[Kingma \& Welling(2014)Kingma and Welling]{kingma2014stochastic}
Diederik~P Kingma and Max Welling.
\newblock Stochastic gradient vb and the variational auto-encoder.
\newblock In \emph{Second International Conference on Learning Representations,
  ICLR}, volume~19, 2014.

\bibitem[Kingma \& Welling(2019)Kingma and Welling]{kingma2019introduction}
Diederik~P Kingma and Max Welling.
\newblock An introduction to variational autoencoders.
\newblock \emph{arXiv preprint arXiv:1906.02691}, 2019.

\bibitem[Kingma \& Dhariwal(2018)Kingma and Dhariwal]{kingma2018glow}
Durk~P Kingma and Prafulla Dhariwal.
\newblock Glow: Generative flow with invertible 1x1 convolutions.
\newblock In \emph{Advances in Neural Information Processing Systems}, pp.\
  10215--10224, 2018.

\bibitem[Kingma et~al.(2016)Kingma, Salimans, Jozefowicz, Chen, Sutskever, and
  Welling]{kingma2016improved}
Durk~P Kingma, Tim Salimans, Rafal Jozefowicz, Xi~Chen, Ilya Sutskever, and Max
  Welling.
\newblock Improved variational inference with inverse autoregressive flow.
\newblock In \emph{Advances in neural information processing systems}, pp.\
  4743--4751, 2016.

\bibitem[Kolesnikov \& Lampert(2017)Kolesnikov and
  Lampert]{kolesnikov2017pixelcnn}
Alexander Kolesnikov and Christoph~H Lampert.
\newblock Pixelcnn models with auxiliary variables for natural image modeling.
\newblock In \emph{International Conference on Machine Learning}, pp.\
  1905--1914. PMLR, 2017.

\bibitem[Maal{\o}e et~al.(2019)Maal{\o}e, Fraccaro, Li{\'e}vin, and
  Winther]{maaloe2019biva}
Lars Maal{\o}e, Marco Fraccaro, Valentin Li{\'e}vin, and Ole Winther.
\newblock Biva: A very deep hierarchy of latent variables for generative
  modeling.
\newblock In \emph{Advances in neural information processing systems}, pp.\
  6548--6558, 2019.

\bibitem[Menick \& Kalchbrenner(2018)Menick and
  Kalchbrenner]{menick2018generating}
Jacob Menick and Nal Kalchbrenner.
\newblock Generating high fidelity images with subscale pixel networks and
  multidimensional upscaling.
\newblock \emph{arXiv preprint arXiv:1812.01608}, 2018.

\bibitem[Oord et~al.(2016)Oord, Dieleman, Zen, Simonyan, Vinyals, Graves,
  Kalchbrenner, Senior, and Kavukcuoglu]{oord2016wavenet}
Aaron van~den Oord, Sander Dieleman, Heiga Zen, Karen Simonyan, Oriol Vinyals,
  Alex Graves, Nal Kalchbrenner, Andrew Senior, and Koray Kavukcuoglu.
\newblock Wavenet: A generative model for raw audio.
\newblock \emph{arXiv preprint arXiv:1609.03499}, 2016.

\bibitem[Papamakarios et~al.(2019)Papamakarios, Nalisnick, Rezende, Mohamed,
  and Lakshminarayanan]{papamakarios2019normalizing}
George Papamakarios, Eric Nalisnick, Danilo~Jimenez Rezende, Shakir Mohamed,
  and Balaji Lakshminarayanan.
\newblock Normalizing flows for probabilistic modeling and inference.
\newblock \emph{arXiv preprint arXiv:1912.02762}, 2019.

\bibitem[Parmar et~al.(2018)Parmar, Vaswani, Uszkoreit, Kaiser, Shazeer, Ku,
  and Tran]{parmar2018image}
Niki Parmar, Ashish Vaswani, Jakob Uszkoreit, {\L}ukasz Kaiser, Noam Shazeer,
  Alexander Ku, and Dustin Tran.
\newblock Image transformer.
\newblock \emph{arXiv preprint arXiv:1802.05751}, 2018.

\bibitem[Radford et~al.(2019)Radford, Wu, Child, Luan, Amodei, and
  Sutskever]{radford2019language}
Alec Radford, Jeffrey Wu, Rewon Child, David Luan, Dario Amodei, and Ilya
  Sutskever.
\newblock Language models are unsupervised multitask learners.
\newblock \emph{OpenAI Blog}, 1\penalty0 (8):\penalty0 9, 2019.

\bibitem[Reed et~al.(2017)Reed, Oord, Kalchbrenner, Colmenarejo, Wang, Belov,
  and De~Freitas]{reed2017parallel}
Scott Reed, A{\"a}ron van~den Oord, Nal Kalchbrenner, Sergio~G{\'o}mez
  Colmenarejo, Ziyu Wang, Dan Belov, and Nando De~Freitas.
\newblock Parallel multiscale autoregressive density estimation.
\newblock \emph{arXiv preprint arXiv:1703.03664}, 2017.

\bibitem[Rezende et~al.(2014)Rezende, Mohamed, and
  Wierstra]{rezende2014stochastic}
Danilo~Jimenez Rezende, Shakir Mohamed, and Daan Wierstra.
\newblock Stochastic backpropagation and approximate inference in deep
  generative models.
\newblock \emph{arXiv preprint arXiv:1401.4082}, 2014.

\bibitem[Salimans \& Kingma(2016)Salimans and Kingma]{salimans2016weight}
Tim Salimans and Durk~P Kingma.
\newblock Weight normalization: A simple reparameterization to accelerate
  training of deep neural networks.
\newblock In \emph{Advances in neural information processing systems}, pp.\
  901--909, 2016.

\bibitem[Salimans et~al.(2017)Salimans, Karpathy, Chen, and
  Kingma]{salimans2017pixelcnn++}
Tim Salimans, Andrej Karpathy, Xi~Chen, and Diederik~P Kingma.
\newblock Pixelcnn++: Improving the pixelcnn with discretized logistic mixture
  likelihood and other modifications.
\newblock \emph{arXiv preprint arXiv:1701.05517}, 2017.

\bibitem[S{\o}nderby et~al.(2016)S{\o}nderby, Raiko, Maal{\o}e, S{\o}nderby,
  and Winther]{sonderby2016ladder}
Casper~Kaae S{\o}nderby, Tapani Raiko, Lars Maal{\o}e, S{\o}ren~Kaae
  S{\o}nderby, and Ole Winther.
\newblock Ladder variational autoencoders.
\newblock In \emph{Advances in neural information processing systems}, pp.\
  3738--3746, 2016.

\bibitem[Strubell et~al.(2019)Strubell, Ganesh, and
  McCallum]{strubell2019energy}
Emma Strubell, Ananya Ganesh, and Andrew McCallum.
\newblock Energy and policy considerations for deep learning in nlp.
\newblock \emph{arXiv preprint arXiv:1906.02243}, 2019.

\bibitem[Uria et~al.(2013)Uria, Murray, and Larochelle]{uria2013rnade}
Benigno Uria, Iain Murray, and Hugo Larochelle.
\newblock Rnade: The real-valued neural autoregressive density-estimator.
\newblock In \emph{Advances in Neural Information Processing Systems}, pp.\
  2175--2183, 2013.

\bibitem[Vahdat \& Kautz(2020)Vahdat and Kautz]{vahdat2020nvae}
Arash Vahdat and Jan Kautz.
\newblock Nvae: A deep hierarchical variational autoencoder.
\newblock \emph{arXiv preprint arXiv:2007.03898}, 2020.

\bibitem[Van~den Oord et~al.(2016)Van~den Oord, Kalchbrenner, Espeholt,
  Vinyals, Graves, et~al.]{van2016conditional}
Aaron Van~den Oord, Nal Kalchbrenner, Lasse Espeholt, Oriol Vinyals, Alex
  Graves, et~al.
\newblock Conditional image generation with pixelcnn decoders.
\newblock In \emph{Advances in neural information processing systems}, pp.\
  4790--4798, 2016.

\bibitem[Zhang et~al.(2019)Zhang, Dauphin, and Ma]{zhang2019fixup}
Hongyi Zhang, Yann~N Dauphin, and Tengyu Ma.
\newblock Fixup initialization: Residual learning without normalization.
\newblock \emph{arXiv preprint arXiv:1901.09321}, 2019.

\bibitem[Zhao et~al.(2017)Zhao, Song, and Ermon]{zhao2017learning}
Shengjia Zhao, Jiaming Song, and Stefano Ermon.
\newblock Learning hierarchical features from generative models.
\newblock \emph{arXiv preprint arXiv:1702.08396}, 2017.

\end{thebibliography}
\bibliographystyle{iclr2021_conference}

\appendix
\section{Appendix}

\subsection{Ablations of architectural components}
\begin{figure}[h]
  \centering
  \includegraphics[width=0.8\linewidth]{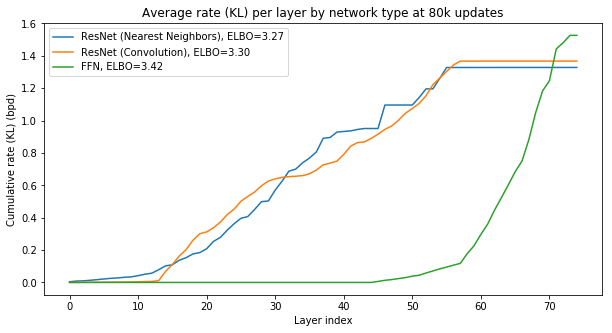}
  \caption{\textbf{Relationship between architecture and posterior collapse.} We visualize the cumulative KL divergence (or ``rate", in bits per dimension) for several different architectures across a 73 layer network on ImageNet-32. When residual connections are removed from the ``res block" in the top-down path (Figure \ref{fig:arch}), the model encodes no information in the first 45 layers of the network and the loss is highest ("FFN"). When a learned convolutional upsampler is used as the ``unpool" layer, the first 13 layers of the network encode no information. When nearest-neighbor upsampling is used, the first layers all encode information, and the loss is the lowest.}
  \label{abl:postcollapse}
\end{figure}

First, we visualize data that suggests upsampling layers and residual connections have an impact on posterior collapse (Figure \ref{abl:postcollapse}). Architectural differences may explain why our VAEs do not need ``free bits" or KL warmups to avoid posterior collapse.

In Table \ref{abl:residualscaling}, we show residual initialization leads to smoother and better training of very deep VAEs. Without residual initialization, very deep VAEs encounter a high number of unstable updates and have higher losses.

\begin{figure}[h]
    \includegraphics[width=.49\textwidth]{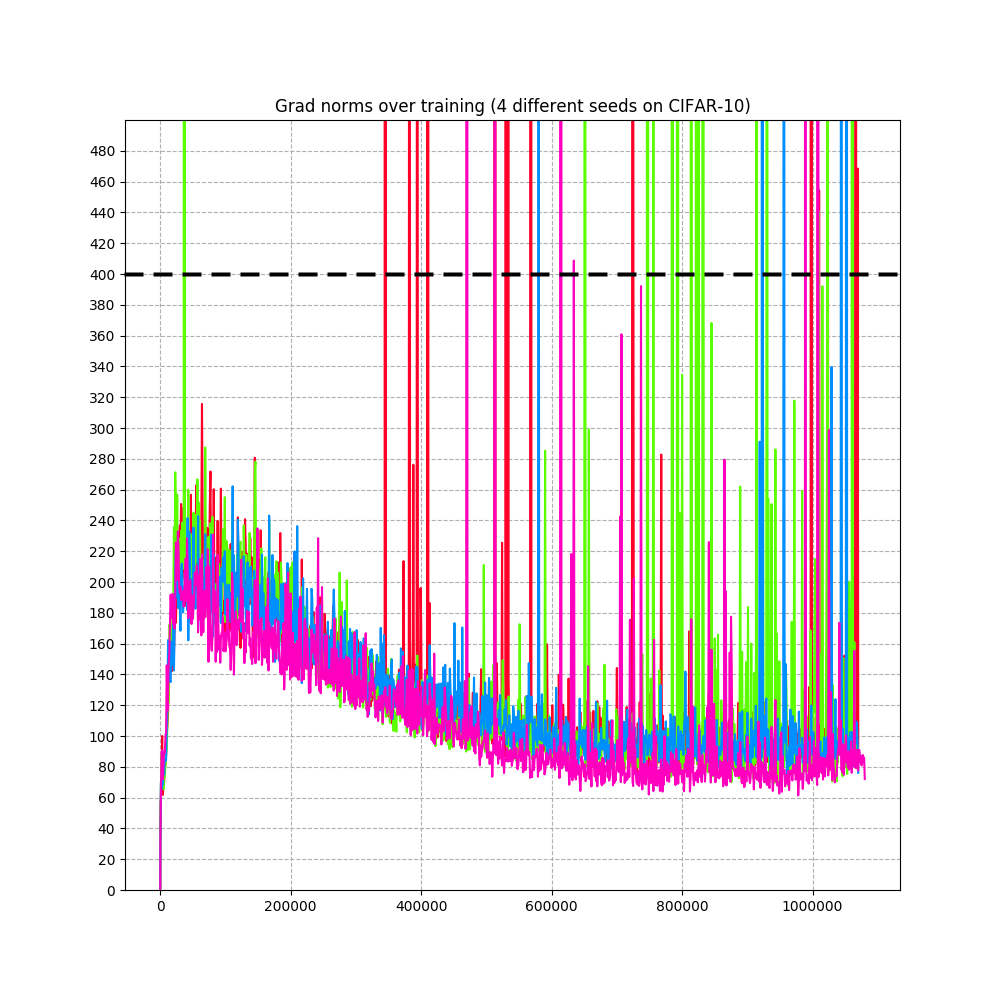}\hfill
    \includegraphics[width=.49\textwidth]{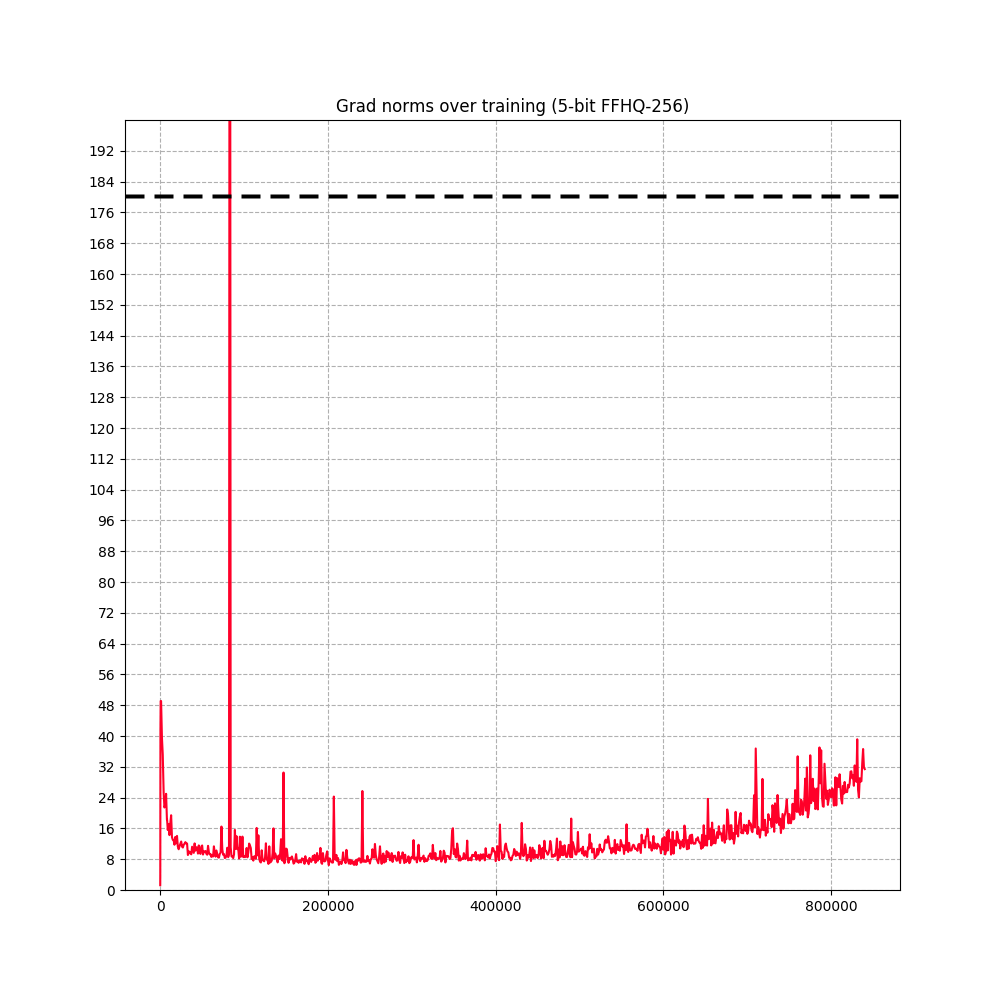}
    \\[\smallskipamount]
    \includegraphics[width=.49\textwidth]{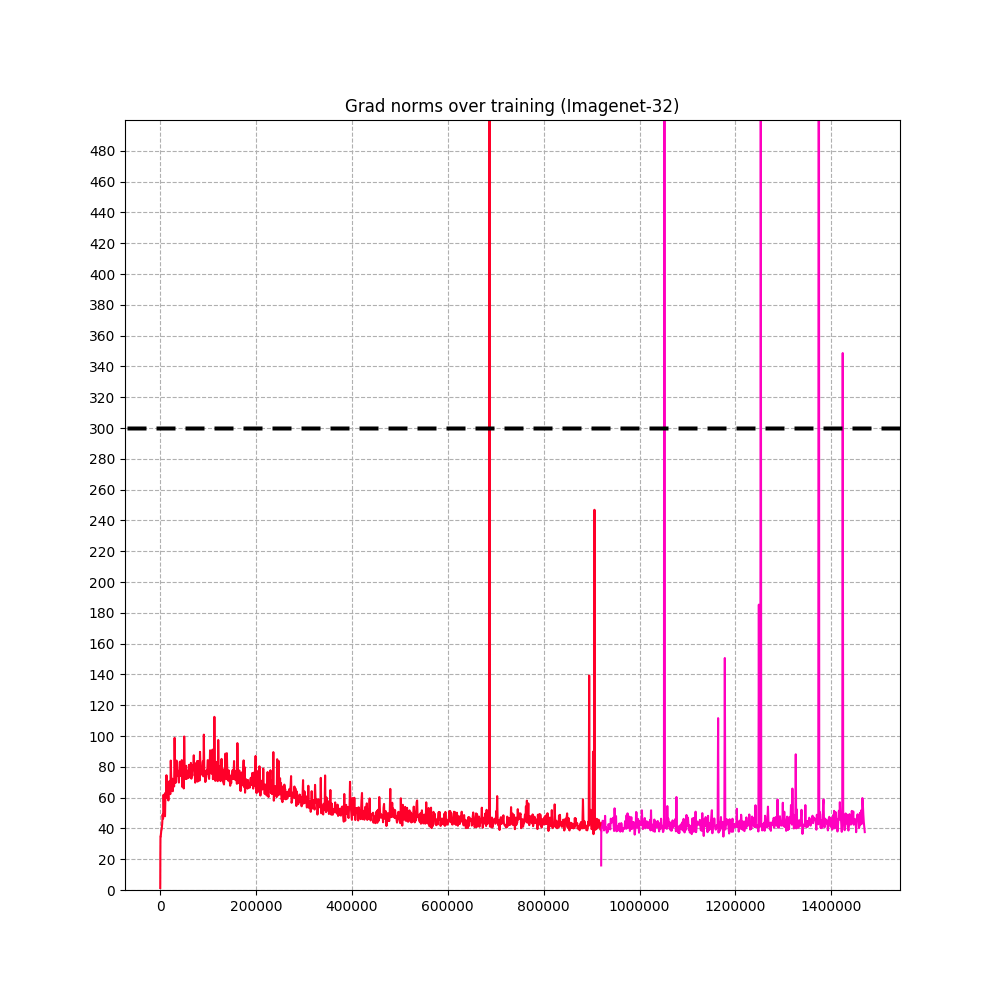}\hfill
    \includegraphics[width=.49\textwidth]{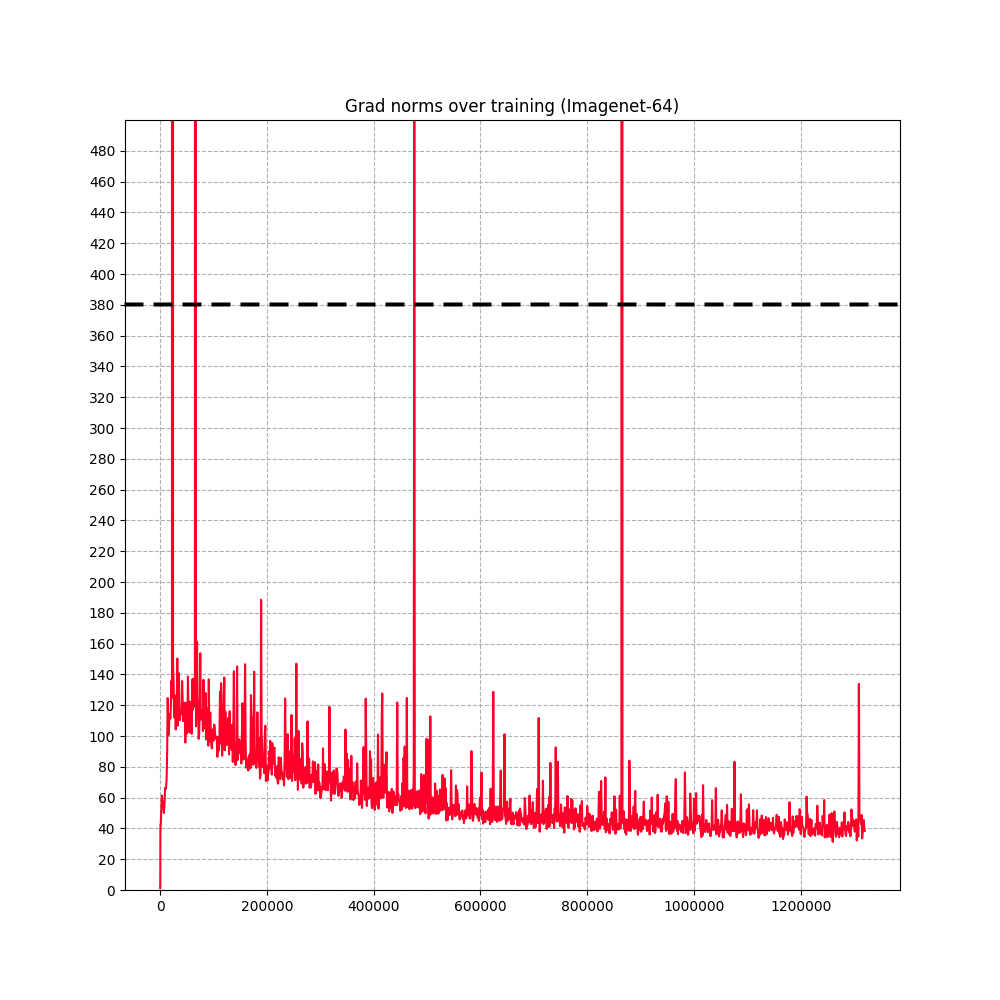}
    \caption{\textbf{Effect of gradient skipping.} We plot the max gradient norm encountered per 500 updates for our best models across datasets. The dashed black line indicates the ``skip threshold", or value above which the update is skipped. We choose a high threshold that affects fewer than 0.01 percent of training updates. Without this skip heuristic, networks will diverge when extreme updates are encountered. These updates can have norm as high as $1e15$.}\label{abl:gradskips}
\end{figure}

In Figure \ref{abl:gradskips}, we show the max gradient norms experienced throughout training, and show that our skipping criterion avoids a small number of updates that would destabilize the network.

\begin{table}[t]
  \caption{\textbf{Effects of scaling residual initialization on very deep VAEs}. We trained networks with varying depths for 80k iterations. Scaling the last layer in the residual block by $\frac{1}{\sqrt{N}}$ results in higher losses for shallower networks, but lower losses and greater stability for deeper networks. The number of updates which are skipped because the gradient norm would destabilize the network is significantly reduced with scaling.}
  \label{abl:residualscaling}
  \centering
  \begin{tabular}{cllll}
    \textbf{Depth} & \textbf{Without scaling} & & \textbf{With scaling}\\
    & Loss & Skipped Updates & Loss & Skipped Updates\\
    \hline
    15 & 2.50 & 13 & 2.51 & 0 \\
    30 & 2.36 & 41 & 2.38 & 1 \\
    45 & 2.31 & 48 & 2.30 & 0 \\
    60 & 2.30 & 76 & 2.29 & 1 \\
    75 & Diverged & - & 2.28 & 0 \\
  \end{tabular}
\end{table}

\begin{figure}[h]
  \centering
  \includegraphics[width=0.8\linewidth]{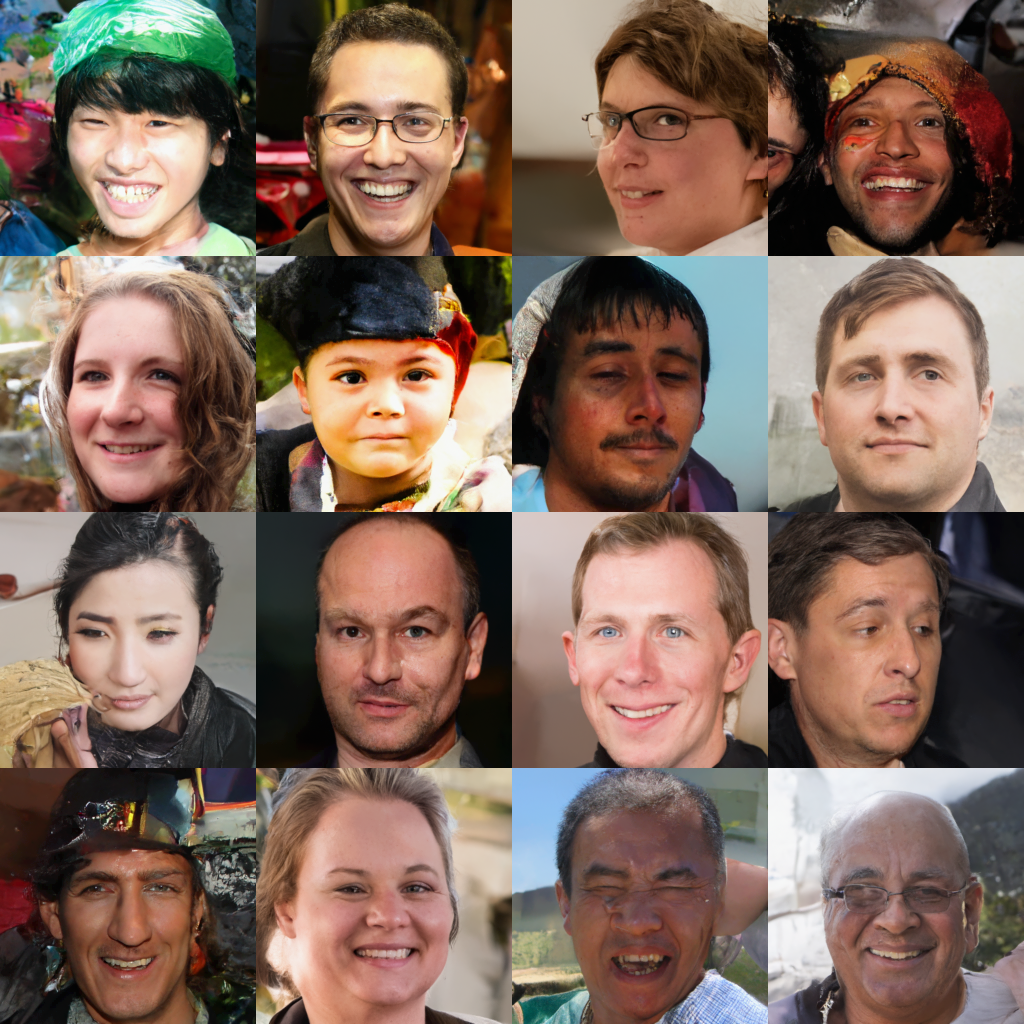}
  \caption{\textbf{Non-cherrypicked, temperature 1.0 samples on FFHQ-256.} Cover images were each cherrypicked from a batch of 16 (unadjusted temperature) samples. Here we show a random batch of 16 images for comparison.}
  \label{samp:ffhq256t1}
\end{figure}

\begin{figure}[h]
  \centering
  \includegraphics[width=0.8\linewidth]{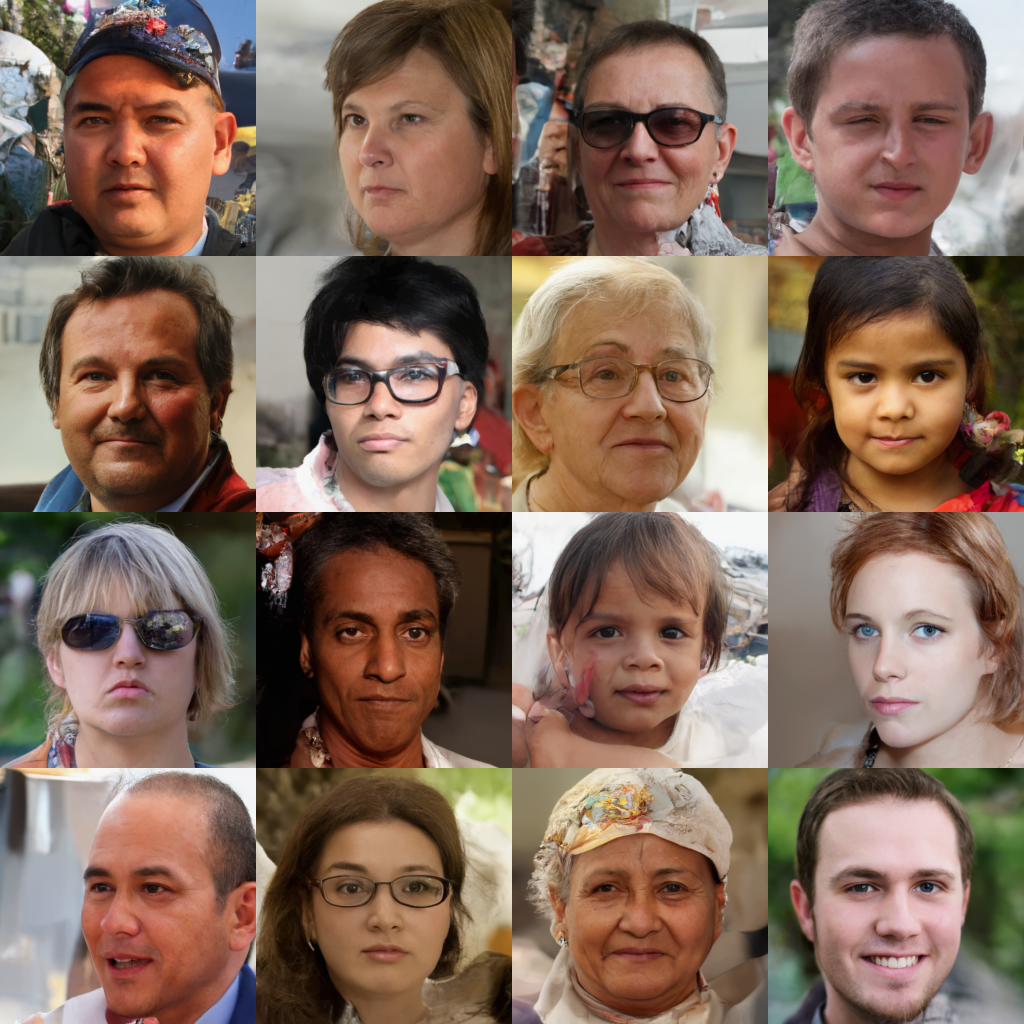}
  \caption{\textbf{Non-cherrypicked, temperature 0.85 samples on FFHQ-256.} Lower temperature samples result in greater regularity in images.}
  \label{samp:ffhq256t60}
\end{figure}

\begin{figure}[h]
  \centering
  \includegraphics[width=0.8\linewidth]{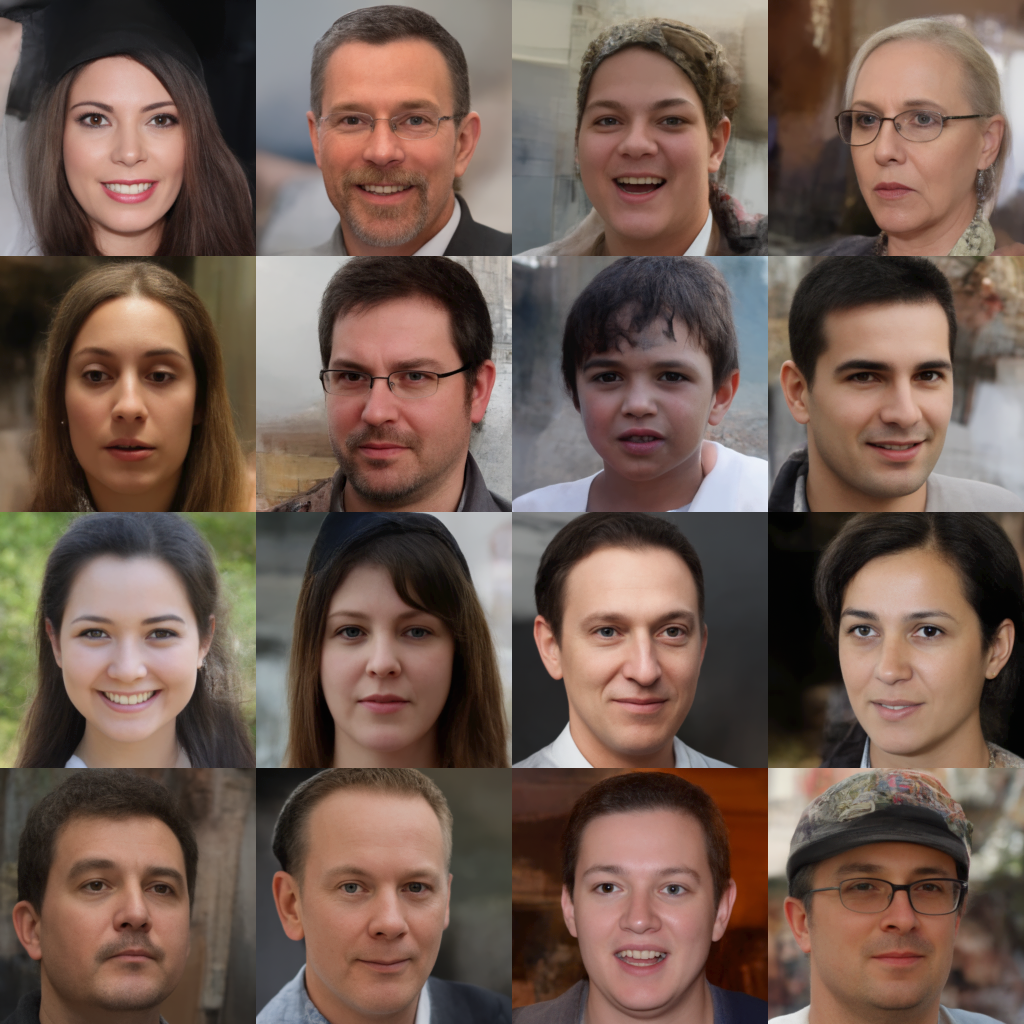}
  \caption{\textbf{Non-cherrypicked, temperature 0.60 samples on FFHQ-256.} We visualize temperature 0.60 samples for comparison with \cite{vahdat2020nvae}}
  \label{samp:ffhq256t60}
\end{figure}

\begin{figure}[h]
  \centering
  \includegraphics[width=1.0\linewidth]{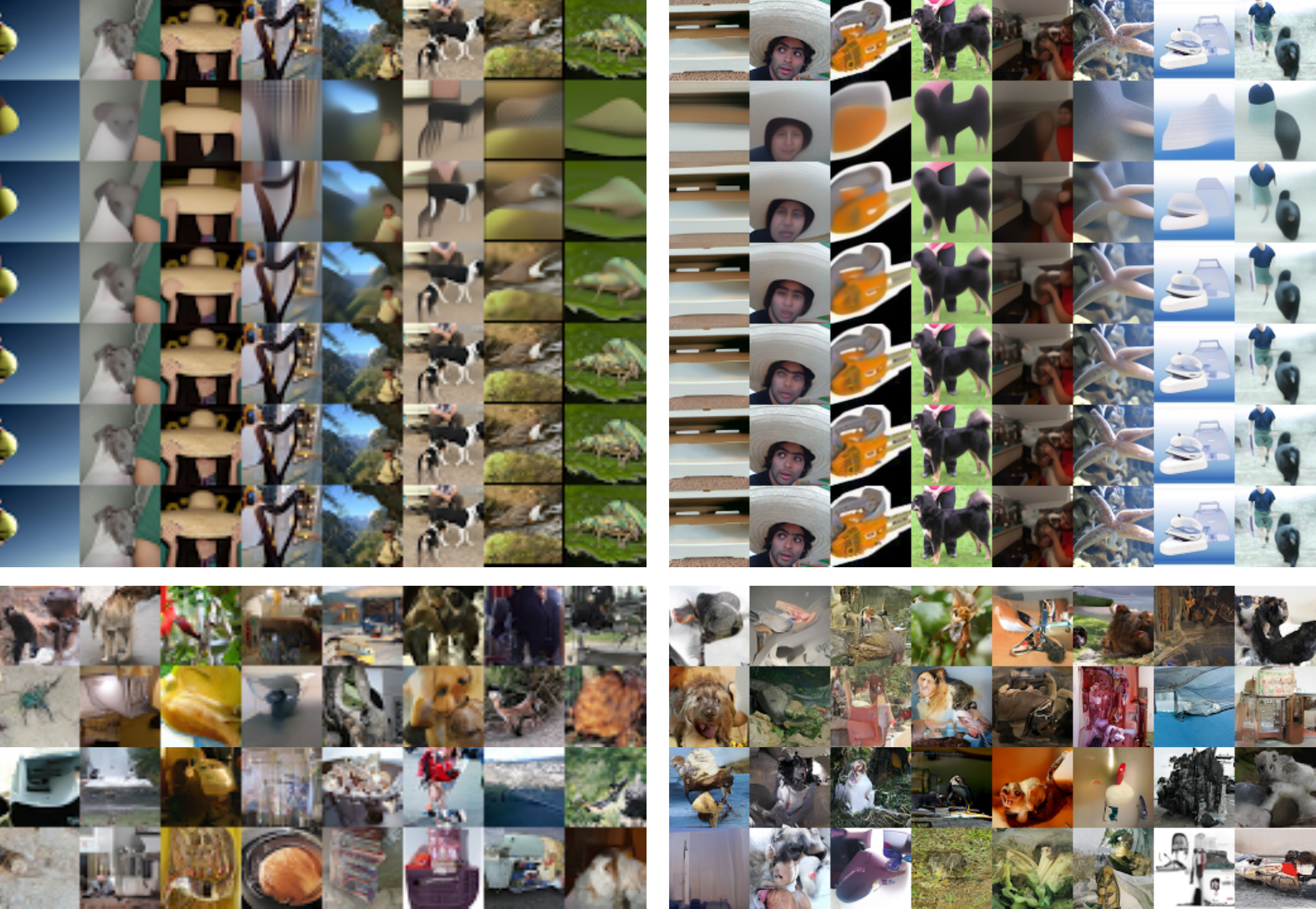}
  \caption{\textbf{ImageNet-32 (left) and ImageNet-64 (right) reconstructions and samples.} Reconstructions of validation images from various stages in the latent hierarchy (top), and unconditional samples from the model at temperature 1.0 (bottom).}
  \label{samp:imagenet}
\end{figure}

\begin{figure}[h]
  \centering
  \includegraphics[width=0.8\linewidth]{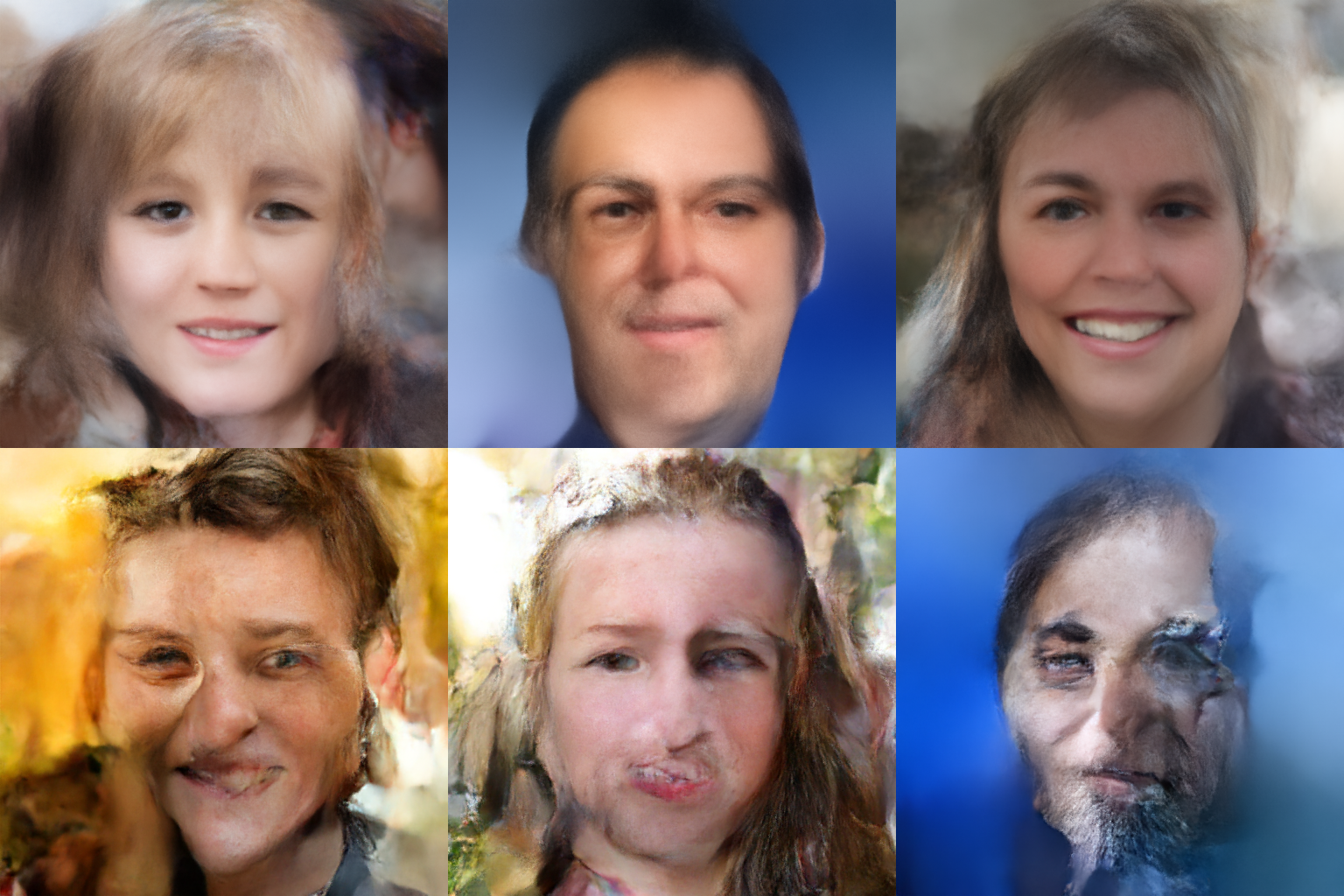}
  \caption{\textbf{FFHQ-1024 samples.} These are generated with reduced temperature (top) and temperature 1.0 (bottom). The model we train has similar capacity to smaller ones we use on 32x32, 64x64, and 256x256 images, and so fails to capture the intricacies of this more complex distribution well. A larger model, trained for longer, may achieve better sample quality.}
  \label{samp:ffhq1024}
\end{figure}

\subsection{Proposition 1: N-layer VAEs generalize autoregressive models when N is the data dimension}

Proposition 1 shows that an autoregressive model with an arbitrary ordering over observed variables in $\vx \in \R^N$ is equivalent to an $N$-layer VAE with an approximate posterior that simply outputs the observed variables in the given order, and a generator that performs the identity function (see Figure \ref{fig:arequiv}).

Without loss of generality, we simplify notation by assuming each vector-valued latent variable $\vz_i$ only has one element, which we write as $z_i \in \R$. We assume a prior and approximate posterior distribution following Equation \ref{topdownpr} and \ref{topdownpo}.

\begin{proof}
Let $q(z_i=x_i| z_{<i}, \vx) = 1$, and $p(x_i=z_i | \vz) = 1$. Then $p(\vz|\vx) = q(\vz|\vx)$, which is well-known to imply equality in the evidence lower bound (ELBO) of Eq. \ref{elbo}. Since $\log q(\vz|\vx) = \log p(\vx|\vz) = 0$, the ELBO becomes $\log p_\theta(\vx) = \log p_\theta(\vz) = \sum_{i=1}^{N} \log p_\theta( z_i | z_{<i}) = \sum_{i=1}^{N} \log p_\theta( x_i | x_{<i})$, which is equivalent to an autoregressive model over the observed variables.
\end{proof}

\subsection{Proposition 2: $N$-layer VAEs are universal approximators of $N$-dimensional latent densities}
Proposition 2 shows that hierarchical VAEs learn depthwise autoregressive flows, and under certain conditions (described in \cite{huang2017learnable}) can express \textit{any} density over latent variables of $N$ dimensions, given enough capacity.

\begin{proof}
We omit full proof, and refer readers to \cite{huang2017learnable,papamakarios2019normalizing}, where universality is established for autoregressive flows. Here we only note that the prior and approximate posterior in an $N$-layer VAE are autoregressive flows: Let $p_\theta(\vz)$ be the prior distribution. $p_\theta(\vz)$ can be written using the reparameterization trick as a deterministic function of noise $\epsilon$ drawn from a known base density $p_N$: $p_\theta(\vz) = p_N(\mathbf{\epsilon}) \left| \mathrm{det} \frac{\partial f(\mathbf{\epsilon}, \theta)}{\partial \mathbf{\epsilon}} \right|$, where $f$ is a neural network that implements the factorization in Eq. \ref{topdownpr}. Since $f$ is autoregressive and its Jacobian is lower triangular, $p_\theta(\vz)$ can approximate any $p(\vz)$ that fits the criteria in \cite{huang2017learnable}. The same logic applies to $q_\phi(\vz | \vx)$ and $p(\vz | \vx)$. It should be noted that this result depends on $f$ being able to implement the inverse CDF of an arbitrary probability density, and so using Gaussian distributions will restrict the densities the VAE can express in practice. This is a limitation of our architecture that we nevertheless adopt since we hypothesize depth, not the elementwise density, is the more important factor. More discussion on this subject, and options for removing this restriction, are described in \cite{huang2017learnable} and \cite{huang2018neural}, and we defer studying more expressive elementwise densities to future work.
\end{proof}

\subsection{A note on Inverse Autoregressive Flow}
\label{iaf}
Inverse autoregressive flows (IAF, \cite{kingma2016improved}) and are similar to very deep VAEs in that they are universal approximators of posterior distributions in VAEs, even with just a \textit{single} layer and sufficiently expressive univariate density \citep{huang2018neural}.

There are several practical differences between IAFs and deep hierarchical VAEs, however, which can result in qualitatively very different behavior. First, the masked autoregressive components in IAF build statistical dependencies \textit{spatially}, whereas a very deep hierarchical VAE builds dependencies \textit{depthwise}, and these inductive biases may better suit different domains. Additionally, IAFs spend an equal amount of computation and parameters on each variable. In contrast, a deep VAE can specify a structure, like a hierarchy of global-to-local variables, which have different computational and modeling capacities for each stage. For images, these differences may result in qualitatively different behavior, and it is not clear whether a single layer IAF can readily learn the sort of rich hierarchical decomposition of images that appear with very deep VAEs.

Nevertheless, the two techniques are complementary -- IAF was introduced in a deep hierarchical VAE \citep{kingma2016improved}, in fact, and it is likely that introducing IAF into our architecture (as in \cite{,vahdat2020nvae}) would improve performance.

\subsection{A note on Learning Hierarchical Features}
\label{zhao}
The work of \cite{zhao2017learning} may appear to contradict our work, by suggesting that additional layers in hierarchical VAEs \textit{do not} lead to additional expressivity, based off their finding that Gibbs sampling from the last stochastic layer is sufficient to recover the data. For high dimensional data like images, however, the last stochastic layer may have many thousands of variables, and Gibbs sampling may take unacceptably long to converge. A hierarchy of latent variables as in our model allows efficient and tractable sampling from this distribution. Additionally, assumptions regarding global maximization of the ELBO may not apply in practice. Nevertheless, we think further clarifying these contradictory statements would be useful future work.

\subsection{Broader Impact}

Broadly speaking, any generative model will reflect the biases of the datasets they are trained on. If deployed without careful consideration, generative models (including but not limited to VAEs) trained on research datasets like ImageNet, CIFAR-10, and FFHQ may inadvertently cause harm by propagating or otherwise reinforcing harmful biases in the dataset. Further work is required to improve and debias research benchmark datasets to mitigate this source of negative impact.

Some VAEs are distinguished from other generative models by their fast synthesis of new data examples. Generative models with fast synthesis can allow for realtime synthesis of high dimensional data, such as music, speech, and video. These models could be used to augment human creativity and lead to a number of helpful applications in real-time media applications. Such models could also be used for compression, which could assist in delivering content to bandwidth-constrained regions of the world. They can also be used for spreading disinformation, generally making it less possible to distinguish real from generated data. An additional potential harm is that fast, high quality synthesis of data could end up economically displacing individuals who rely upon creative work, such as musicians, visual artists, and more.

VAEs also are distinguished by their usage of latent variables. Generative models with useful latent variables could have positive impacts in scientific domains, where density estimation could lead to novel insights about chemical, physical, or biological data. Latent variable representations of data could also be helpful in efforts to debias, interpret, or otherwise increase understandibility of models and their representations.

\begin{table}[h]
  \caption{\textbf{Key hyperparameters for experiments}. We detail here the main hyperparameters used in training. FFHQ-1024 has reduced hidden size for higher resolutions; see code for details.}
  \label{hparams}
  \centering
  \begin{tabular}{cccccc}
    \textbf{Parameter} & CIFAR-10 & ImageNet-32 & ImageNet-64 & FFHQ-256 & FFHQ-1024\\
    \hline
    Num layers & 45 & 78 & 75 & 62 & 72\\
    Hidden size & 384 & 512 & 512 & 512 & Varies\\
    Bottleneck size & 96 & 128 & 128 & 128 & Varies\\
    Latent dim per layer & 16 & 16 & 16 & 16 & 16\\
    Batch size & 32 & 256 & 128 & 32 & 32\\
    Learning rate & 0.0002 & 0.00015 & 0.00015 & 0.00015 & 0.00007\\
    Optimizer & Adam & Adam & Adam & Adam & Adam\\
    Skip threshold & 400 & 300 & 380 & 180 & 500\\
    Weight Decay & 0.01 & 0.0 & 0.0 & 0.0 & 0.0\\
    EMA rate & 0.0002 & 0.00015 & 0.00015 & 0.00015 & 0.00015\\
    Training iterations & 1.1M & 1.7M & 1.6M & 1.7M & 1.7M\\
    GPUs & 2 x V100 & 32 x V100 & 32 x V100 & 32 x V100 & 32 x V100\\
    Training time & 6 days & 2.5 weeks & 2.5 weeks & 2.5 weeks & 2.5 weeks\\
    Parameters & 39M & 119M & 125M & 115M & 115M\\
  \end{tabular}
\end{table}

\end{document}

\section{Appendix}

Here we describe details of our network architecture and training procedure. Then we provide hyperparameters for all experimental results.

\subsection{Network architecture}
Our architecture is a modified version of a top-down \cite{sonderby2016ladder,kingma2016improved} hierarchical VAE. It has a deterministic \textit{encoder}, a stochastic \textit{inference network}, and a combined generator/prior network that we call the \textit{generator}. All three networks operate at a number of spatial resolutions, progressing from high resolution at the bottom of the network (usually the data resolution $D$, 32x32 or 64x64) to low resolution at the top of the network (typically 1x1). A visual representation of the network is available in Figure 2 (left).

\subsubsection{Encoder}
The encoder takes in an image and outputs a series of activations $\{h_{D \times D}, h_{\frac{D}{k} \times \frac{D}{k}}, ..., h_{1 \times 1}\}$, one for each resolution in the network. The number of resolutions, as well as the number of \textit{blocks} per resolution, is a hyperparameter, which we specify like the following example:

\begin{verbatim}
32x10,16x10,8x10,4x10,1x10
\end{verbatim}

This means that there are 10 blocks at resolution 32x32, followed by 10 at 16x16, and so forth. When changing resolutions, we use a single $k \times k$ convolution to downsample, where $k$ is the degree of downsampling required. For a given resolution $r$, $h_{r \times r}$ is equal to the activations after all blocks of that resolution.

Each block is a bottleneck residual block \cite{he2016deep} with two 3x3 convolutions in the middle, no normalization, and the GELU \cite{hendrycks2016gaussian} nonlinearity. In pseudocode:

\begin{verbatim}
def block_encoder(h):
  h_next = c1(gelu(h))
  h_next = c2(gelu(h_next))
  h_next = c3(gelu(h_next))
  return h + c4(gelu(h_next))
\end{verbatim}

Where c1 and c4 are 1x1 convolutions to/from $w, \frac{1}{4}w$ and c2, c3 are 3x3 convolutions to/from $\frac{1}{4}w, \frac{1}{4}w$.  We zero-initialize the weights of c4, and zero-initialize the bias of all convolutions. Also, at 1x1 resolution we only used 1x1 convolutions instead of any 3x3 convolutions.

\subsubsection{Inference and Generator networks}
The stochastic inference network and generator network share many parameters, and so we describe them here jointly. The network starts with a 1x1 resolution ``tentative output'' $\hat{x}_{1 \times 1}$, and processes it with a number of blocks at each resolution before upsampling to the next resolution, and proceeding until $\hat{x}_{D \times D}$ is created and fed to the loss. The blocks per resolution is a hyperparameter we specify like the following example:

\begin{verbatim}
1x1,4x2,8x5,16x10,32x11
\end{verbatim}

Each block combines the approximate posterior and prior in the following way, described in pseudocode:

\begin{verbatim}
def block_decoder(xhat):
  posterior_distribution = posterior_net([xhat, h])
  prior_distribution = posterior_net([xhat, feedforward_1(xhat)])
  z = sample_from_gaussian(posterior_distribution)
  xhat = feedforward_2(xhat + linear(z))
  kl = kl_divergence(posterior_distribution, prior_distribution)
  return xhat, kl
\end{verbatim}

Here, $h$ refers to the encoder activations $h_{r \times r}$ of the current resolution.

The posterior network is exactly the same as the bottleneck block, except instead of the last residual connection, it outputs the mean and log standard deviation of a Gaussian distribution. The feedforward networks each consist of two (non-residual) 3x3 convolutional layers interspersed with the GELU nonlinearity. Each of these convolutions has a group size of 4 to reduce parameter count, and has zero-initialized biases.

It should be noted that we re-use posterior network parameters to define the prior, which reduces parameter count and which we found sometimes improved performance. This re-using of parameters was first introduced by \citet{tomczak2017vae}, where the authors directly parameterized "pseudo-inputs" which were processed by the posterior to create a prior distribution. Our formulation above can be viewed as an adaptation to deep hierarchical VAEs, where a parametric network outputs the "pseudo-input" for each distribution. We only have the network output one "pseudo-input", but a mixture distribution can also be created by creating multiple such inputs.

\subsubsection{Input and Output}
The raw image input is normalized according to the mean and standard deviation of the training set, then processed with a 3x3 convolution with $w$ number of output filters.

The output layer uses the discretized mixture of logistics loss introduced in \citet{salimans2017pixelcnn++} and used in other work on VAEs \cite{maaloe2019biva}.

\subsection{Training procedure}
\label{train_procedure}
All networks were either optimized with Adam \cite{kingma2014adam} or AdamW \cite{loshchilov2017decoupled} (with weight decay coefficient specified below), on the per-subpixel ELBO loss. 

We adjust the loss slightly in the following way. We found that when the KL divergence between posterior and prior was trained directly, the networks could fluctuate to extreme values. So we instead train the posterior against the standard normal distribution, and train the prior to predict the posterior values. Toward the end of training, when parameter values are less likely to change, we restart training with the KL between the posterior and prior.

Additionally, we found that sometimes updates would be of very high or NaN gradient norm. Although we found that limiting the precision of Gaussian distributions by, for instance, using a softplus activation on the log standard deviation \cite{maaloe2019biva} could address this, we found that simply skipping updates of gradients with very large magnitude also works and allows for smooth optimization to convergence. We typically chose to skip gradient updates where the norm (as calculated by PyTorch's torch.nn.utils.clip\_grad\_norm\_ function) was above 1000.

On CIFAR-10, ImageNet-32 and ImageNet-64, we used a validation set of 5000 randomly sampled from the training dataset in order to tune hyperparameters. We evaluated with one sample on the test set, instead of using many importance weighted samples as introduced in \citet{burda2015importance} and used in other work \cite{maaloe2019biva,ho2019flow++}, as we found it did not provide a significant gain in our models. We suspect this may be due to the distributions our network learns being very peaky.

We used Polyak averaging of the training weights for evaluation, with coefficient 0.999.

\subsection{Hyperparameters for Table 1}
All of our networks evaluated in Table 1 had the following hyperparameters, unless otherwise stated:

\begin{verbatim}
Width: 384
Encoder blocks: 32x7,16x7,8x7,4x7,1x7
Decoder blocks: 1x6,4x6,8x6,16x6,32x6
Learning rate: 2e-4
Batch size: 32
Gaussian dimension per layer: 16
Weight decay: 0.0
\end{verbatim}

They were trained for 200 epochs, corresponding to approximately 1 day on one V100, with the validation loss reported after 200 epochs. The networks had approximately 40M parameters.

\subsection{Hyperparameters for Table 2}

\subsubsection{CIFAR-10}
The network we report had the following hyperparameters:

\begin{verbatim}
Width: 384
Encoder blocks: 32x10,16x10,8x10,4x10,1x10
Decoder blocks: 1x1,4x2,8x5,16x10,32x11
Learning rate: 2e-4
Batch size: 32
Gaussian dimension per layer: 16
Weight decay: 0.005
\end{verbatim}

It was trained for 650 epochs, corresponding to approximately 2-3 days on one V100, then we trained with the posterior-prior KL (see \ref{train_procedure}) for another 10 epochs before evaluating on the test set. It had 42M total parameters.

\subsubsection{ImageNet-32}
The network we report had the following hyperparameters:

\begin{verbatim}
Width: 768
Encoder blocks: 32x7,16x7,8x7,4x7,1x7
Decoder blocks: 1x3,4x3,8x2,16x10,32x8
Learning rate: 1.5e-4
Batch size: 128
Gaussian dimension per layer: 64
Weight decay: 0.0
\end{verbatim}

It was trained for 70 epochs, then another 100 epochs with the posterior-prior KL, amounting to approximately 7 days on 8 V100 GPUs. It had 112M parameters.

\subsubsection{ImageNet-64}
The network we report had the following hyperparameters:

\begin{verbatim}
Width: 1024
Encoder blocks: 64x6,32x6,16x6,8x6,4x6,1x6
Decoder blocks: 1x5,4x5,8x5,16x5,32x5,64x3
Learning rate: 1.5e-4
Batch size: 32
Gaussian dimension per layer: 64
Weight decay: 0.0
\end{verbatim}

We trained for 25 epochs, then trained for another 25 with the posterior-prior KL and a reduced learning rate of 5e-5. This took approximately 14 days on 8 V100 GPUs. It had 258M parameters.

\small

\bibliography{refs}{}
\bibliographystyle{plainnat}